\documentclass[11pt, a4paper]{lumia}

\usepackage[sort&compress]{natbib}
\usepackage{xspace}

\theoremstyle{plain}

\newtheorem*{proposition*}{Proposition}

\theoremstyle{definition}

\theoremstyle{definition}

\def\eqref#1{equation~\ref{#1}}

\usepackage{graphicx}           
\usepackage{tikz}               
\usepackage[edges]{forest}      

\usepackage{url}                
\usepackage{xurl}               

\usepackage{array}              
\usepackage{longtable}          
\usepackage{multirow}           
\usepackage{makecell}           
\usepackage{ragged2e}           

\usepackage{mathtools}          
\usepackage{nicefrac}           

\usepackage{algorithm}          
\usepackage{algorithmicx}       
\usepackage{algpseudocode}      
\usepackage{listings}           

\usepackage{subcaption}         
\usepackage{wrapfig}            
\usepackage[export]{adjustbox}  

\usepackage{changes}            
\usepackage{xspace}             
\usepackage[normalem]{ulem}     
\usepackage{CJKutf8}            

\usepackage[tikz]{bclogo}       
\usepackage[framemethod=tikz]{mdframed} 

\usepackage{lipsum}             
\usepackage{tocloft}            
\usepackage{afterpage}          
\usepackage{bbding}             
\usepackage{epigraph}           
\usepackage{minitoc}            
\usepackage{multicol}           
\usepackage{textgreek}          

\newcolumntype{P}[1]{>{\RaggedRight\arraybackslash}p{#1}}

\definecolor{uclablue}{RGB}{39, 116, 174}
\definecolor{bigaired}{RGB}{156, 0, 0}
\definecolor{myblue}{HTML}{598BE7}
\definecolor{mildblue}{RGB}{31,119,180}
\definecolor{sectionblue}{RGB}{70, 130, 180}
\definecolor{methodblue}{RGB}{0, 150, 136}
\definecolor{bgblue}{RGB}{245,243,253}
\definecolor{ttblue}{RGB}{91,194,224}
\definecolor{mygreen}{rgb}{0.64, 0.56, 0.88}
\definecolor{myyellow}{rgb}{0.68, 0.6, 0.1}
\definecolor{fancygreen}{rgb}{0.33, 0.68, 0.20}
\definecolor{salmon}{rgb}{0.94, 0.52, 0.49}
\definecolor{tablegreen}{rgb}{0.82, 0.94, 0.75}
\definecolor{tableblue}{rgb}{0.81, 0.90, 0.94}
\definecolor{tablered}{rgb}{0.97, 0.85, 0.85}
\definecolor{tableorange}{rgb}{0.96, 0.85, 0.81}
\definecolor{myorange}{rgb}{1.0, 0.49, 0.0}
\definecolor{tlgreen}{rgb}{0.33, 0.68, 0.20}
\definecolor{darkgreen}{RGB}{0,100,0}
\definecolor{darkred}{RGB}{200, 0, 0}
\definecolor{customyellow}{HTML}{FFFACD}
\definecolor{refinegreen}{RGB}{0, 128, 75}
\definecolor{scoregreen}{RGB}{34, 139, 34}
\definecolor{hidden-blue}{RGB}{194,232,247}
\definecolor{hidden-black}{RGB}{20,68,106}
\definecolor{yes}{HTML}{C6EFCE}
\definecolor{no}{HTML}{FFC7CE}
\definecolor{partial}{HTML}{FFEB9C}
\definecolor{external}{HTML}{D9E1F2}
\definecolor{hdr}{HTML}{F2F2F2}
\definecolor{GRPOrow}{gray}{0.96}
\definecolor{FlowRLrow}{RGB}{225,236,255}
\definecolor{FlowBlue}{RGB}{80,120,210}
\definecolor{GRPOGray}{gray}{0.35}

\hypersetup{
    colorlinks=true,
    citecolor=uclablue,
    linkcolor=bigaired,
    urlcolor=darkblue
}

\setlist[itemize]{leftmargin=20pt, noitemsep, topsep=0pt}

\NewDocumentCommand{\kaiyan}{mO{}}{\textcolor{purple}{\textsuperscript{\textit{kaiyan}}\textsf{\textbf{\small[#1]}}}}
\NewDocumentCommand{\yuxin}{mO{}}{\textcolor{cyan}{\textsuperscript{\textit{yuxin}}\textsf{\textbf{\small[#1]}}}}
\NewDocumentCommand{\bx}{mO{}}{\textcolor{green}{\textsuperscript{\textit{bx}}\textsf{\textbf{\small[#1]}}}}
\NewDocumentCommand{\at}{mO{}}{\textcolor{red}{\textsuperscript{\textit{AT}}\textsf{\textbf{\small[#1]}}}}
\NewDocumentCommand{\re}{mO{}}{\textcolor{blue}{\textsuperscript{\textit{RE}}\textsf{\textbf{\small[#1]}}}}
\NewDocumentCommand{\ybsun}{mO{}}{\textcolor{magenta}{\textsuperscript{\textit{youbang}}\textsf{\textbf{\small[#1]}}}}
\NewDocumentCommand{\runze}{mO{}}{\textcolor{orange}{\textsuperscript{\textit{runze}}\textsf{\textbf{\small[#1]}}}}
\NewDocumentCommand{\add}{mO{}}{\textcolor{darkgreen}{\textsuperscript{\textit{Maybe Consider Discuss}}\textsf{\textbf{[#1]}}}}

\newcommand{\cmark}{\textcolor{darkgreen}{\boldmath$\checkmark$}}
\newcommand{\xmark}{\textcolor{darkred}{\boldmath$\times$}}

\newenvironment{itemize*}%
 {\leftmargini=10pt\begin{itemize}%
  \setlength{\itemsep}{0pt}%
  \setlength{\parskip}{0pt}%
  }%
 {\end{itemize}}

\newenvironment{enumerate*}%
 {\begin{enumerate}%
  \setlength{\itemsep}{0pt}%
  \setlength{\parskip}{0pt}}%
 {\end{enumerate}}

\newcommand{\cellstatus}[1]{%
  \begingroup
  \StrTrim{#1}[\statusval]%
  \IfStrEq{\statusval}{Yes}{\cellcolor{yes}\cmark}{}%
  \IfStrEq{\statusval}{No}{\cellcolor{no}\xmark}{}%
  \IfBeginWith{\statusval}{Yes (}{\cellcolor{yes}\cmark~\textit{\statusval\unskip}}{}%
  \IfStrEq{\statusval}{Partial}{\cellcolor{partial}\textbf{Partial}}{}%
  \IfStrEq{\statusval}{External}{\cellcolor{external}\textbf{External}}{}%
  \endgroup
}

\newtcolorbox{myboxi}[1][]{
  breakable,
  title=#1,
  colback=red!5,
  colbacktitle=red!5,
  coltitle=black,
  fonttitle=\bfseries,
  bottomrule=0pt,
  toprule=0pt,
  leftrule=2pt,
  rightrule=2pt,
  titlerule=0pt,
  arc=0pt,
  outer arc=0pt,
  colframe=red,
}

\newtcolorbox{myboxnote}[1][]{
  breakable,
  title=#1,
  colback=orange!0,
  colbacktitle=orange!0,
  coltitle=black,
  fonttitle=\bfseries,
  bottomrule=0pt,
  toprule=0pt,
  leftrule=2pt,
  rightrule=2pt,
  titlerule=0pt,
  arc=0pt,
  outer arc=0pt,
  colframe=orange,
}

\newtcolorbox{myboxii}[1][]{
  breakable,
  freelance,
  title=#1,
  colback=white,
  colbacktitle=white,
  coltitle=black,
  fonttitle=\bfseries,
  bottomrule=0pt,
  boxrule=0pt,
  colframe=white,
  overlay unbroken and first={
  \draw[red!75!black,line width=3pt]
    ([xshift=5pt]frame.north west) --
    (frame.north west) --
    (frame.south west);
  \draw[red!75!black,line width=3pt]
    ([xshift=-5pt]frame.north east) --
    (frame.north east) --
    (frame.south east);
  },
  overlay unbroken app={
  \draw[red!75!black,line width=3pt,line cap=rect]
    (frame.south west) --
    ([xshift=5pt]frame.south west);
  \draw[red!75!black,line width=3pt,line cap=rect]
    (frame.south east) --
    ([xshift=-5pt]frame.south east);
  },
  overlay middle and last={
  \draw[red!75!black,line width=3pt]
    (frame.north west) --
    (frame.south west);
  \draw[red!75!black,line width=3pt]
    (frame.north east) --
    (frame.south east);
  },
  overlay last app={
  \draw[red!75!black,line width=3pt,line cap=rect]
    (frame.south west) --
    ([xshift=5pt]frame.south west);
  \draw[red!75!black,line width=3pt,line cap=rect]
    (frame.south east) --
    ([xshift=-5pt]frame.south east);
  },
}

\tcbset{
  takeawaysbox/.style={
    title=Takeaways,
    colback=lightblue!80,
    colframe=black,
    fonttitle=\bfseries\small,
    coltitle=white,
    colbacktitle=black,
    enhanced,
    attach boxed title to top left={xshift=2.5mm,yshift=-2.5mm},
    boxed title style={rounded corners, size=small, colframe=black, colback=black},
    width=\linewidth,
    arc=3.5mm
  }
}

\mdfdefinestyle{mystyle}{%
  rightline=true,
  innerleftmargin=10,
  innerrightmargin=10,
  outerlinewidth=3pt,
  topline=false,
  rightline=true,
  bottomline=false,
  skipabove=\topsep,
  skipbelow=\topsep
}

\tikzset{%
    every node/.style={font=\tiny},
    parent/.style =          {align=center,text width=2cm,rounded corners=3pt, line width=0.3mm, fill=gray!10,draw=gray!80},
    child/.style =           {align=center,text width=2.0cm,rounded corners=3pt, fill=blue!10,draw=blue!80,line width=0.3mm},
    grandchild/.style =      {align=center,text width=2cm,rounded corners=3pt},
    greatgrandchild/.style = {align=center,text width=1.5cm,rounded corners=3pt},
    greatgrandchild2/.style = {align=center,text width=1.5cm,rounded corners=3pt},
    referenceblock/.style =  {align=center,text width=1.5cm,rounded corners=2pt},
    pretrain/.style =           {align=center,text width=2.0cm,rounded corners=3pt, fill=blue!10,draw=blue!80,line width=0.3mm},
    pretrain_work/.style =           {align=center, text width=8.5cm,rounded corners=3pt, fill=blue!10,draw=blue!0,line width=0.3mm},
    template/.style =           {align=center,text width=2.0cm,rounded corners=3pt, fill=red!10,draw=red!80,line width=0.3mm},
    template_work/.style =           {align=center,text width=8.5cm,rounded corners=3pt, fill=red!10,draw=red!0,line width=0.3mm},
    answer/.style =           {align=center,text width=2.0cm,rounded corners=3pt, fill= cyan!10,draw= cyan!80,line width=0.3mm},
    answer_work/.style =           {align=center,text width=8.5cm,rounded corners=3pt, fill= cyan!10,draw= cyan!0,line width=0.3mm},
    multiple/.style =           {align=center,text width=2.0cm,rounded corners=3pt, fill= orange!10,draw= orange!80,line width=0.3mm},
    multiple_work/.style =           {align=center,text width=8.5cm,rounded corners=3pt, fill= orange!10,draw= orange!0,line width=0.3mm},
    tuning/.style =           {align=center,text width=2.0cm,rounded corners=3pt, fill= magenta!10,draw= magenta!80,line width=0.3mm},
    tuning_work/.style =           {align=center,text width=8.5cm,rounded corners=3pt, fill= magenta!10,draw= magenta!0,line width=0.3mm},
}

\newcommand{\lstbg}[3][0pt]{{\fboxsep#1\colorbox{#2}{\strut #3}}}

\lstdefinelanguage{diff}{
  basicstyle=\ttfamily\small,
  morecomment=[f][\lstbg{red!20}]-,
  morecomment=[f][\lstbg{green!20}]+,
}

\lstdefinelanguage{diffpython}{
  language=diff,
  morekeywords={def, if, else, for, while, return, import, from, as, class, with, try, except, finally, raise, lambda, and, or, not, in, is, None, True, False},
  morecomment=[l]{\#},
  morestring=[b]",
  morestring=[b]',
}

\setheadertext{LUMIA Lab}

\correspondingemail{\emailicon\ weirubinn@gmail.com, lin.zhouhan@gmail.com \quad $*$ Equal Contribution \quad $^\ddagger$ Corresponding Author.}

\githublink{https://github.com/LUMIA-Group/MLPMemory}\huggingfacelink{https://huggingface.co/collections/Rubin-Wei/mlpmemory}

\renewcommand{\today}{2025-10-23}

\title{MLP Memory: A Retriever-Pretrained Memory for Large Language Models}
\setheadertitle{MLP Memory: A Retriever-Pretrained Memory for Large Language Models}

\author{%
  Rubin Wei$^{1,*}$, Jiaqi Cao$^{1,4*}$, Jiarui Wang$^{1}$, Jushi Kai$^{1}$, Qipeng Guo$^{2}$, Bowen Zhou$^{2,3}$, Zhouhan Lin$^{1,2\ddagger}$\\
  $^1$LUMIA Lab, School of Artificial Intelligence, Shanghai Jiao Tong University \\
  $^2$Shanghai AI Laboratory  \quad
  $^3$Department of Electronic Engineering, Tsinghua University \quad
  $^4$SJTU Paris Elite Institute of Technology \\

}

\begin{document}

\begin{abstract}
Modern approaches to enhancing Large Language Models' factual accuracy and knowledge utilization face a fundamental trade-off: non-parametric retrieval-augmented generation (RAG) provides flexible access to external knowledge but suffers from high inference latency and shallow integration, while parametric fine-tuning methods like LoRA risk catastrophic forgetting and degraded general capabilities. In this work, we propose MLP Memory, a lightweight parametric module that learns to internalize retrieval patterns without explicit document access. By pretraining an MLP to imitate a $k$NN retriever's behavior on the entire pretraining dataset, we create a differentiable memory component that captures the benefits of retrieval-based knowledge access in a fully parametric form. Our architecture integrates this pretrained MLP Memory with Transformer decoders through simple probability interpolation, yielding 17.5\% and 24.1\% scaling gains on WikiText-103 and Web datasets, respectively. It further achieves 12.3\% relative improvement on five question-answering benchmarks and 5.2 points absolute gain across nine general NLP tasks, while reducing hallucinations by up to 10 points on HaluEval. Moreover, MLP Memory delivers 2.5$\times$ faster inference than RAG with superior accuracy. Our findings show that learning retrieval patterns parametrically bridges the gap between efficient inference and effective knowledge access, offering a practical alternative to both RAG and fine-tuning approaches.
\end{abstract}

\maketitle


\begin{figure}[h]
\centering
\includegraphics[width=\textwidth]{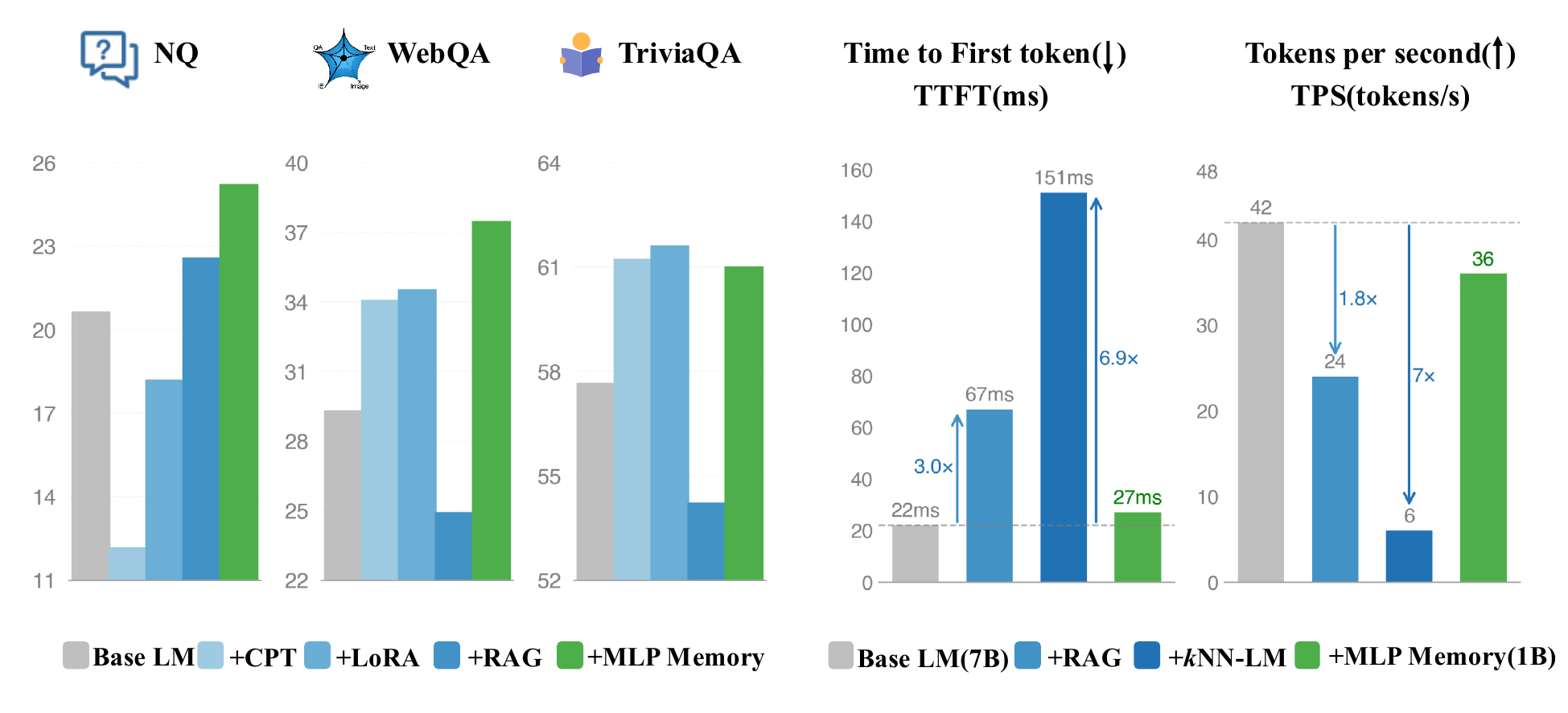}
\caption{Performance and efficiency comparison. \textbf{Left}: accuracy across three QA benchmarks. MLP Memory consistently outperforms the base model, surpassing both parametric methods (CPT, LoRA) and non-parametric retrieval (RAG). \textbf{Right}: inference efficiency, measured by time to first token (TTFT, $\downarrow$ lower is better) and tokens per second (TPS, $\uparrow$ higher is better). RAG results are shown for top-5 retrieval. $k$NN-LM is accelerated via dimension reduction (4096$\rightarrow$256), and both RAG and $k$NN-LM use the Wikipedia-2021 retrieval corpus. MLP Memory uses 1B parameters.}
\label{fig:figure_1}
\end{figure}

\section{Introduction}
\label{intro}

Decoder-only architectures such as GPT~\citep{gpt3}, LLaMA~\citep{llama3}, Qwen~\citep{qwen2.5}, and DeepSeek~\citep{deepseekv3} have achieved remarkable success in various tasks, including open-ended text generation~\citep{gpt4}, code completion~\citep{gpt-code}, image synthesis~\citep{GPT-image}, and multimodal reasoning~\citep{llava}. However, despite their impressive capabilities, these models often struggle with effective knowledge utilization, producing responses that may be fluent but fail to accurately leverage the factual information encoded in their parameters.

\begin{figure}[t]
\centering
\includegraphics[width=\textwidth]{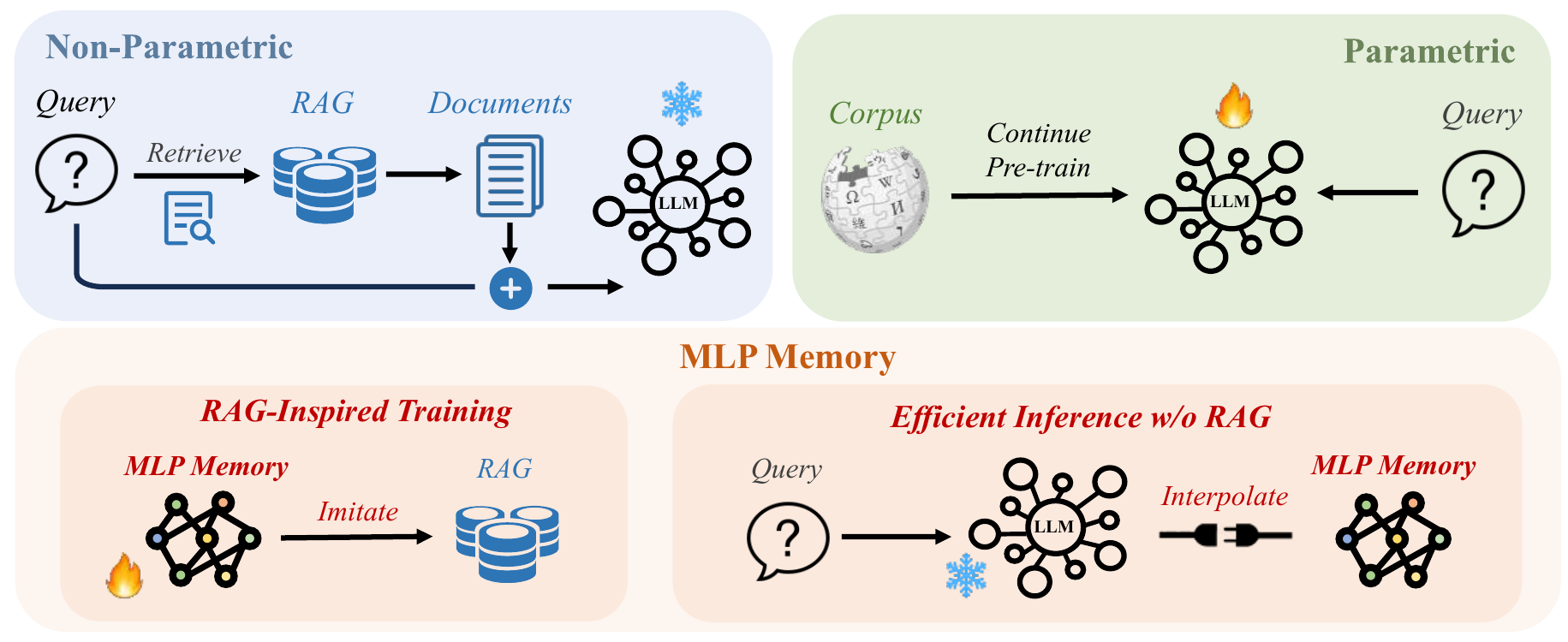}
\caption{Approaches to enhance factual accuracy and knowledge utilization. Top left: Non-parametric RAG provides flexible knowledge access but suffers from high latency. Top right: Parametric fine-tuning risks catastrophic forgetting. Bottom: MLP Memory learns retrieval patterns during training (left) and enables efficient inference without explicit retrieval (right).}
\label{fig:different_method}
\end{figure}

Current approaches to enhance knowledge utilization in LLMs face significant trade-offs. Retrieval-augmented generation (RAG) methods ~\citep{RAG1, RAG2, RAG3, RAG-atlas} dynamically fetch relevant documents to ground model outputs, providing flexible access to external knowledge sources. However, these non-parametric approaches introduce substantial inference latency through expensive nearest-neighbor searches and longer context from retrieved documents. They also suffer from shallow integration with the base model, as the retrieval component remains isolated from the LLM's computational graph. Conversely, parametric adaptation methods such as continued pre-training (CPT) and LoRA~\citep{LoRA} directly modify model weights to incorporate domain-specific knowledge. While computationally efficient at inference time, these approaches risk catastrophic forgetting of previously learned capabilities and often degrade performance on general tasks, requiring careful task-specific tuning that limits their broader applicability. Figure \ref{fig:different_method} illustrates how our approach differs fundamentally from both non-parametric retrieval methods and parametric adaptation approaches.


\begin{figure}[t]
\centering
\includegraphics[width=\textwidth]{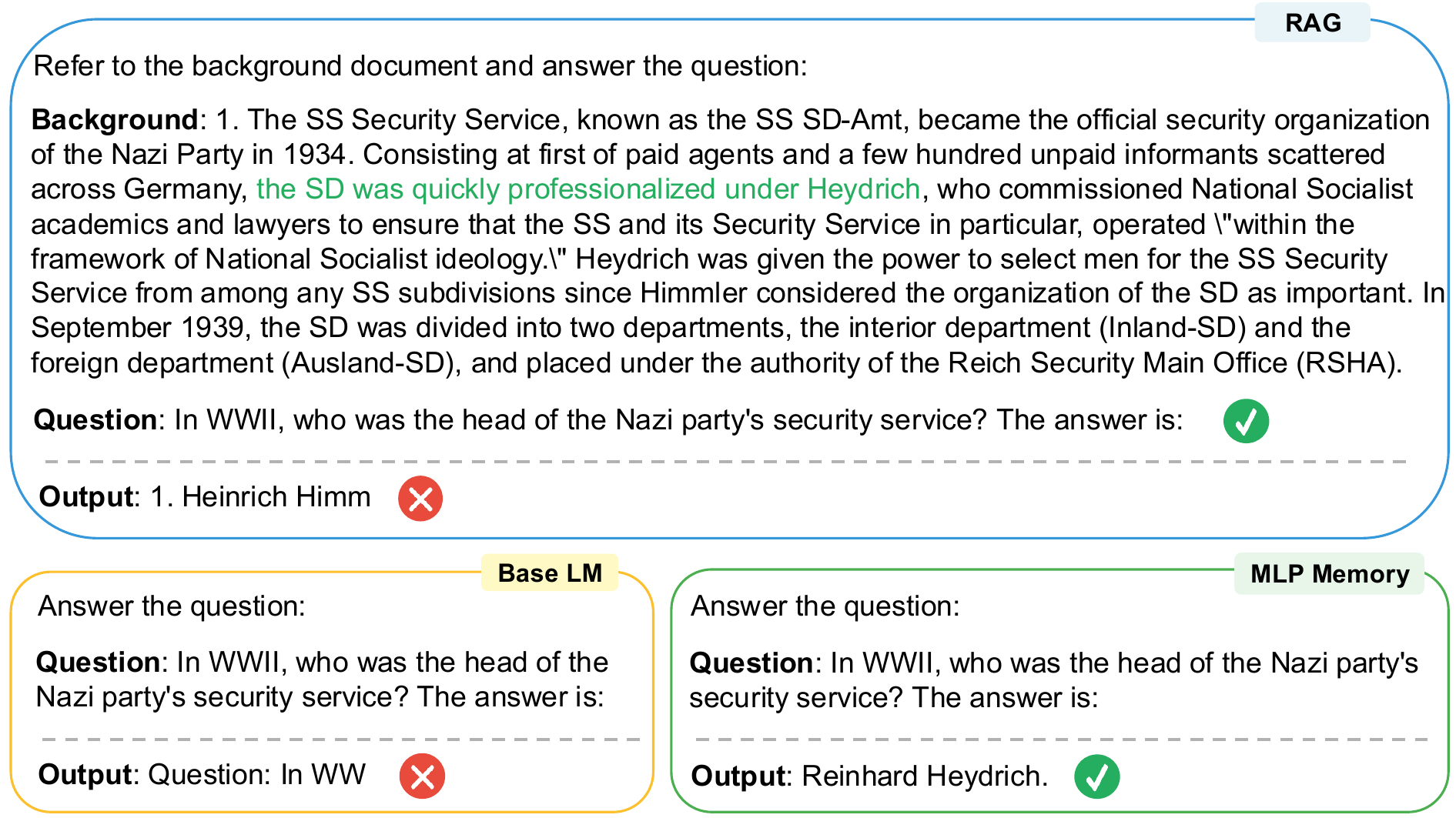}
\caption{Comparison of model outputs on a factual question. Despite retrieving relevant documents with correct information (highlighted in green), RAG is misled by contextual distractors and produces an incorrect answer. MLP Memory generates the correct answer without explicit retrieval.}
\label{fig:figure_3}
\end{figure}

In contrast to decoder-only LLMs, neuroscience research reveals a lateralized human brain where language processing is dominated by the left hemisphere while memory formation occurs in the hippocampus~\citep{Split-Brain, EthicalBrain, hippocampus}. This insight has inspired memory-augmented models in machine learning. Early approaches like Memory Networks~\citep{MemoryNetworks} enabled read/write operations on external memory, while Sparse Access Memory introduced differentiable memory access schemes. However, these were task-specific with limited general applicability. In the LLM era, methods such as Memory Transformers~\citep{MemoryTransformer} incorporate trainable memory tokens for global context, while AutoCompressors~\citep{AutoCompressor} compress long contexts into summary vectors. Nevertheless, these memory tokens primarily function as working memory supplements for context extension rather than long-term memory capable of retaining information from the entire training corpus.

In this work, we propose an external memory for LLM that is pretrained to mimic a retriever on the entire pretraining dataset. Specifically, following the RAG setting in $k$NN-LM~\citep{knnlm}, this memory learns to map the LLM hidden state at a certain step to a vocabulary distribution matching the output of the $k$NN retriever. During inference, the LLM's native output is interpolated with the retriever-pretrained output from the external memory. Our resulting architecture, illustrated in Figure \ref{fig:overview}, consists of a transformer decoder and an external MLP memory, each pretrained separately with different pretraining tasks. For our pretrained external memory, we aim to achieve the following features simultaneously:
\begin{enumerate}[label=\arabic*), labelindent=1.0em, leftmargin=*, align=right, labelsep=0.5em]
    \item \textbf{End-to-end differentiability.} Unlike the non-parametric nature of retrievers, our MLP memory is fully parameterized and allows gradieat flow during training. This enables end-to-end joint optimization of the entire model architecture.
    \item \textbf{Highly compressible memory.} The MLP memory compresses large datastores (e.g., 40TB for 5B tokens in $k$NN-LM) into a compact parametric form (e.g., 4GB for 1B parameters storing 5B tokens), facilitating efficient deployment without performance degradation.
    \item \textbf{Low inference-time latency.} MLP memory eliminates costly retrieval operations, achieving 2.5$\times$ faster inference than RAG methods and 5.6$\times$ faster inference than kNN-LM when using a 5B-token retrieval corpus. Crucially, unlike retrieval-based approaches, our method's inference speed remains constant regardless of the retrieval corpus size.

    \item \textbf{Scalable, general knowledge memorization, covering the whole training set.} MLP memory is trained on the entire pretraining dataset, not limited to the context level. It supports general knowledge retention and demonstrates stronger scaling behavior than decoder-only models.

    \item \textbf{Long-term memory.} While existing memory tokens serve primarily as working memory by storing local context for immediate use, our MLP memory functions as a long-term repository of generalizable knowledge acquired during the pretraining phase.

\end{enumerate}


Experimental results demonstrate that MLP Memory significantly outperforms existing approaches across multiple dimensions. Our architecture exhibits steeper power-law scaling with model size, achieving 17.5\% and 24.1\% improvement on WikiText-103 and Web datasets compared to decoder-only architectures, while continuing to benefit from additional training without overfitting. On downstream tasks, it achieves average relative improvements of 12.3\% (Mistral-7B) and 7.8\% (Llama2-7B) on five QA benchmarks, with WebQA showing exceptional gains (37.45\% vs. 29.28\% baseline). On nine general NLP tasks, it delivers a 5.2 points absolute improvement. MLP Memory also substantially reduces hallucinations on HaluEval, with accuracy improvements of 9.68, 10.08, and 2.14 points on dialogue, QA, and summarization tasks respectively. Most notably, it achieves 2.5$\times$ faster time-to-first-token than RAG and 5.6$\times$ faster than $k$NN-LM, while maintaining constant inference speed regardless of corpus size, unlike retrieval methods whose latency scales with data size. Figure~\ref{fig:figure_1} illustrates MLP Memory's performance gains and inference efficiency over baselines and Figure~\ref{fig:figure_3} demonstrates a case where MLP Memory correctly answers factual questions while RAG fails despite retrieving correct information. These results confirm that parametric compression of retrieval patterns offers a more efficient and effective alternative to explicit retrieval.

\section{Preliminary: $k$-nearest neighbors language model}
\label{Preliminary}

The $k$NN-LM~\citep{knnlm} augments a pre-trained LM by interpolating its parametric distribution with a non-parametric distribution from nearest neighbor retrieval. Given context $c_t = (w_1, ...,w_{t-1})$, and $w_t$ denotes the next token. The next-token probability is:
\begin{equation}
    p(w_t \mid c_t) = \lambda \, p_{kNN}(w_t \mid c_t) + (1-\lambda) \, p_{LM}(w_t \mid c_t),
\end{equation}
where $\lambda \in [0,1]$ is the interpolation parameter, $p_{LM}$ is the LM's distribution, and $p_{kNN}$ is retrieval-based distribution.

\textbf{Datastore} Constructed via a forward pass over a corpus, the datastore consists of key-value pairs $(k_t, v_t)$ where $k_t = f(c_t)$ encodes context $c_t$ using LM representations, and $v_t$ is the next token $w_t$:
\begin{equation}
    (\mathcal{K}, \mathcal{V}) = \left\{ \left( f(c_t), w_t \right) \mid (c_t, w_t) \in \mathcal{D} \right\}.
\end{equation}

\textbf{Inference} The LM encodes context $c$ into query $f(c)$ and retrieves $k$-nearest neighbors $\mathcal{N}$ from $(\mathcal{K}, \mathcal{V})$ using distance metric $d(\cdot,\cdot)$ (typically squared $L^2$). The non-parametric distribution is:
\begin{equation}
    p_{kNN}(y \mid c) \propto \sum_{(k_i, v_i) \in \mathcal{N}} \mathbb{I}_{y=v_i} \, \exp(-d(k_i, f(c))).
\end{equation}

While $k$NN-LM improves predictions through explicit memory, it suffers from substantial storage requirements and high-latency retrieval. For instance, the Wikitext-103 datastore requires nearly 500 GB of storage even for the GPT2-small model~\citep{Efficient-knnlm}. These limitations motivate our MLP Memory, a compact parametric model pretrained to approximate the retrieval function: given a query embedding, it directly outputs a $k$NN-like next token distribution, thereby eliminating both the substantial storage requirements and high-latency retrieval.

\section{MLP Memory}
\label{sec:method}
\begin{figure}[t]
\centering
\includegraphics[width=\textwidth]{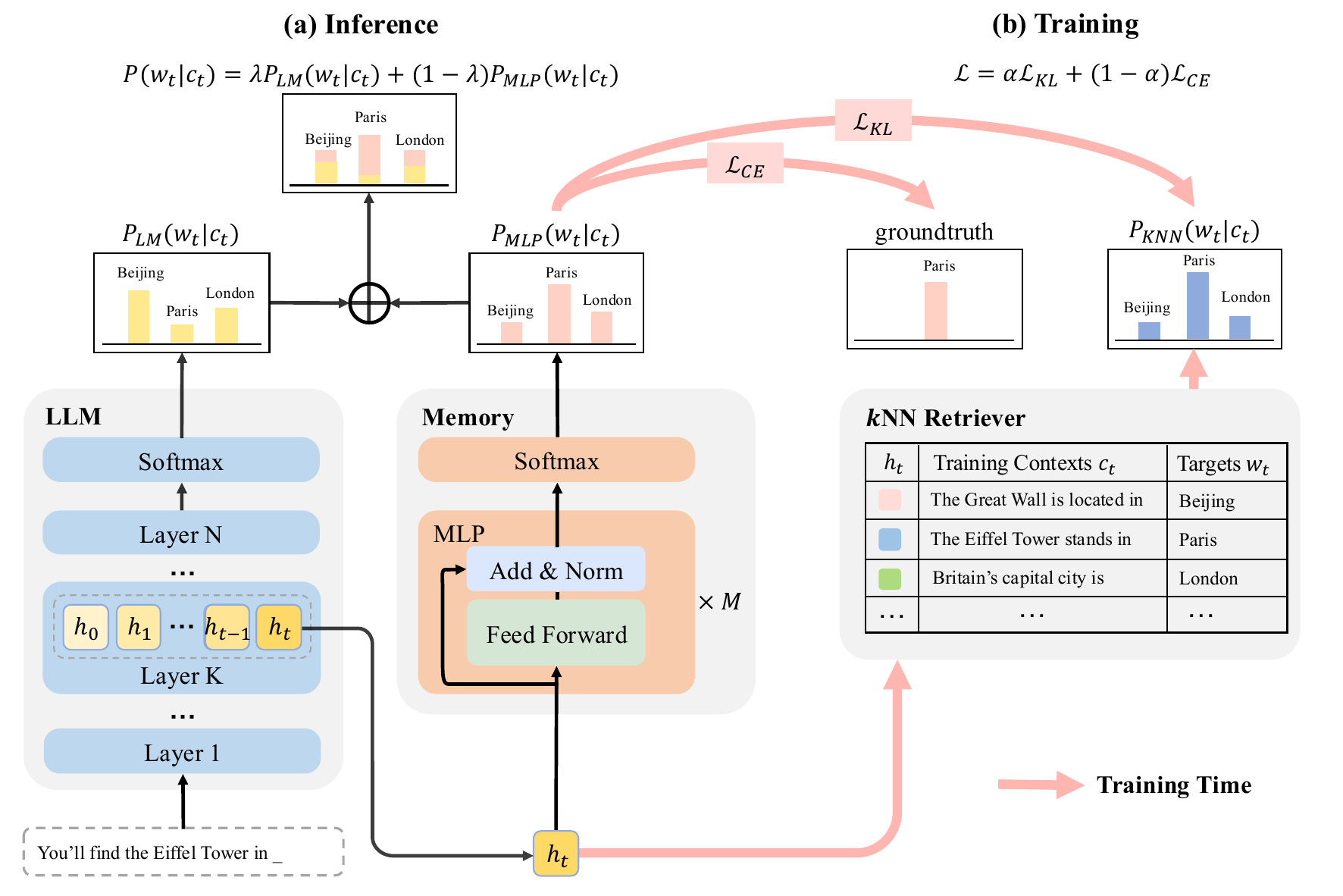}
\caption{Overview of MLP Memory architecture. (a) Inference: MLP Memory processes context representations from a specific LLM layer, generating token probabilities that are interpolated with LLM outputs for final predictions. (b) Training: MLP Memory learns to imitate retriever behavior using LLM representations as input and distributions generated by $k$NN retrievers as targets, optimized through a hybrid objective.}
\label{fig:overview}
\end{figure}

In this section, we present MLP Memory, a lightweight parametric module that learns to internalize retrieval patterns without explicit document access. Our approach consists of three key components: a stack of MLPs that processes hidden representations without token-mixing operations (Section \ref{section.3.1}), a specialized pre-training procedure that enables the MLP to mimic non-parametric retrieval distributions (Section \ref{section.3.2}), and an efficient inference mechanism for deployment (Section \ref{sec:inference}). As illustrated in Figure \ref{fig:overview}, MLP Memory first learns to mimic non-parametric retrieval distributions during pre-training (Figure \ref{fig:overview}(b)), then seamlessly integrates with the language model during inference (Figure \ref{fig:overview}(a)), eliminating both the storage requirements of large datastores and the computational cost of nearest neighbor search.

\subsection{Architecture}
\label{section.3.1}
Our MLP Memory learns to mimic non-parametric retrieval by mapping query embeddings to $k$NN distributions. Given query $q = f(c)$ from context $c$, the MLP directly predicts $p_{kNN}(y|c)$ without neighbor search, transforming discrete retrieval into a differentiable mapping $\mathcal{M}: \mathbb{R}^d \rightarrow \mathbb{R}^{|V|}$, where $d$ is the embedding dimension and $| V |$ is the vocabulary size.

In designing the memory module, we observe from Section \ref{Preliminary} that the retriever imitation task processes a single-vector representation without requiring token-mixing operations. Recent studies~\citep{FFN-KV} have identified that FFN layers function as key-value memories, suggesting that MLPs play a specialized role in knowledge memorization within LLMs. Based on these insights, we propose pretraining an all-MLP memory that effectively functions as a non-parametric retriever, as illustrated in Figure \ref{fig:overview}.

The MLP Memory takes hidden representations $f(c)$ from the pretrained LM as input and is trained to predict the corresponding $k$NN distribution $p_{kNN}(y|c)$ as its target. Once trained, the MLP's output distribution is interpolated with the LM's parametric distribution during inference, following the same interpolation scheme as $k$NN-LM but without requiring datastore access or neighbor search.


\subsection{Training}
\label{section.3.2}
The training procedure for MLP Memory consists of two primary stages: constructing supervision signals from non-parametric retrieval distributions, and optimizing the MLP to mimic these distributions through a carefully designed loss function.

\textbf{Data Construction\quad}
To generate supervision for training MLP Memory, we leverage the datastore construction process described in Section \ref{Preliminary}. We build the datastore $(\mathcal{K}, \mathcal{V})$ through a forward pass over the training corpus, storing context representations and their corresponding next tokens. For each training example $(c_t, w_t) \in \mathcal{D}$, we compute the non-parametric distribution $p_{kNN}(y|c_t)$ by retrieving $k$-nearest neighbors from the datastore. To prevent trivial self-retrieval that would contaminate the learning signal, we exclude the query itself from the neighbor set when constructing the target distribution. These embedding-distribution pairs $\{(f(c_t), p_{kNN}(\cdot|c_t))\}$ are precomputed offline and cached for efficient training.

\textbf{Loss Function\quad}
Unlike traditional language modeling with single-label targets, $k$NN distributions capture the diversity of plausible continuations by encoding multiple valid next tokens weighted by their contextual similarity.
Our ablation studies in Section~\ref{ablation_study} demonstrate that a hybrid objective combining two complementary losses yields optimal performance. Our approach centers on minimizing the Kullback-Leibler divergence~\citep{KL} between MLP Memory's output distribution and the cached $k$NN distributions:
\begin{align}
\mathcal{L}_{KL}(c_t) = \text{KL}(p_{kNN}(\cdot|c_t) \parallel p_{MLP}(\cdot|c_t))
\end{align}
This encourages the memory module to match the full probability distribution rather than merely predicting the most likely token.
To prevent excessive deviation from the underlying corpus distribution, we integrate a complementary Cross-Entropy loss~\citep{CEloss}:
\begin{align}
\mathcal{L}_{CE}(c_t) = -\log p_{MLP}(w_t|c_t)
\end{align}
The final training objective balances these two components through a hyperparameter $\alpha$:
\begin{align}
\mathcal{L}(c_t) = \alpha \cdot \mathcal{L}_{KL}(c_t) + (1-\alpha) \cdot \mathcal{L}_{CE}(c_t)
\end{align}
The KL term encourages learning distributional patterns while the CE term ensures accurate ground-truth prediction, preventing the overfitting that occurs with cross-entropy alone.

\subsection{Inference}
\label{sec:inference}

Once trained, MLP Memory integrates with the base language model through simple probability interpolation.
During inference, MLP Memory processes hidden representations from the language model $\mathcal{M}_{\text{LM}}$ and produces a distribution that is interpolated with the LM's output:
\begin{align}
p_{final}(w_t|c_t) = \lambda \cdot p_{MLP}(w_t|c_t) + (1-\lambda) \cdot p_{LM}(w_t|c_t)
\end{align}
where $\lambda \in [0,1]$ controls the influence of retrieval-based knowledge.

Unlike retrieval-augmented approaches that require nearest neighbor search and extended context processing, MLP Memory requires only a single forward pass through a lightweight all-MLP architecture.
As demonstrated in Figure~\ref{fig:figure_1}, our method achieves 2.5$\times$ faster time-to-first-token than RAG (top-5) and 5.6$\times$ faster than $k$NN-LM, despite $k$NN-LM employing dimension reduction from 4096 to 256 for acceleration.
For tokens per second, MLP Memory delivers 1.5$\times$ higher throughput than RAG and 6$\times$ higher than $k$NN-LM, while introducing only 1.2$\times$ overhead relative to the base model.
Crucially, this performance remains constant regardless of retrieval corpus size, unlike retrieval-based methods whose latency scales with datastore size.

\section{Experimental Setup}
\label{sec:experiments_setup}
\textbf{Overview\quad} We conduct comprehensive experiments to evaluate MLP Memory across five critical dimensions. First, we analyze the scaling behavior (\ref{sec.5.1}) of models augmented with MLP memory through power-law fitting. Second, we assess performance on five question-answering benchmarks (\ref{sec.5.2}) to demonstrate that our approach represents a novel form of parametric memory that surpasses both traditional parametric methods (continued pretraining, LoRA) and non-parametric approaches (RAG). Third, we evaluate on fundamental NLP tasks (\ref{general_nlp_task}) to verify that integrating MLP Memory preserves the base model's general capabilities. Fourth, we examine hallucination reduction (\ref{sec.5.4}) on HaluEval to validate our method's effectiveness in improving factual accuracy. Finally, we present an ablation study (\ref{ablation_study}) to analyze design choices such as loss weighting and layer selection.

\textbf{Implementation Details\quad}
We conduct our experiments on 32×A800 80GB GPUs. To demonstrate the generalizability of our approach, we employ two distinct backbone models: Llama-2-7B~\citep{llama2} and Mistral-7B-v0.3~\citep{Mistral-7B}. For scaling law, please refer to Section~\ref{sec.5.1} for detailed settings. For question-answering benchmarks, we build key-value datastores and non-parametric distributions using both models on preprocessed Wikipedia-2021~\citep{RAG-atlas}, and train separate 1B-parameter MLP Memory modules with learning rate 4e-4. The MLP Memory uses 8 layers by default. See Appendix \ref{mlp_architecture} for details. For general NLP tasks, we build datastores using Mistral-7B-v0.3 on a heterogeneous corpus following~\cite{knnlm-limits}, and train the MLP Memory with learning rate 4e-4. For hallucination evaluation, we directly apply the MLP Memory trained from question-answering experiments. All experiments use a training budget equivalent to the computational cost of training a 7B parameter model for 1 epoch. The training hyperparameter $\alpha$ is set to 0.4 across all tasks. The interpolation hyperparameter $\lambda$ is tuned on the validation split of each task following ~\cite{knnlm}, see more details in Appendix \ref{sensitivity_interpolation}.

\textbf{Baselines\quad}
We compare MLP Memory against established methods for improving factual accuracy and knowledge utilization:
\textbf{RAG}, which employs BGE~\citep{bge} as the retrieval model and retrieves top-5 documents to ensure comprehensive context coverage. \textbf{$k$NN-LM}~\citep{knnlm}, configured with interpolation parameter $\lambda=0.1$ and temperature $\tau=10.0$ following~\citep{knnlm-limits}. \textbf{LoRA}~\citep{LoRA}, applied to query, key, value, and MLP layers, with rank adjusted to match the parameter count of our MLP Memory modules. \textbf{Continued Pretraining (CPT)}, which involves further training of all model parameters on the corresponding corpus.

\section{Experimental Results}

\subsection{Scaling law}
\label{sec.5.1}
\paragraph{Setup} We conduct scaling law experiments using standard decoder-only models and our overall model architecture. As baselines, we use four GPT-2~\cite{gpt2} variants with increasing parameter counts: GPT2-small (124M), GPT2-medium (345M), GPT2-large (774M), and GPT2-xl (1.5B). For MLP Memory, we define three configurations: small (124M), medium (335M), and large (774M) that align with the scaling trend of standard architectures. The MLP Memory module is externally integrated with a matching-sized GPT-2 variant, resulting in total parameter counts of approximately 248M, 710M, and 1.5B for the small, medium, and large configurations, respectively. All models are trained on two datasets: WikiText-103~\cite{Wikitext103} (around 100M tokens) and a mixed Web dataset (around 600M tokens). Following ~\cite{Webdataset}, our Web dataset combines diverse knowledge sources relevant to common NLP tasks, including WikiText-103, Amazon Reviews~\cite{AmazonReview}, CC-NEWS~\cite{CC-NEWs}, and IMDB~\cite{IMDB}.

\begin{figure}[t]
  \centering
  \begin{subfigure}[b]{0.32\textwidth} 
    \centering
    \includegraphics[width=\textwidth]{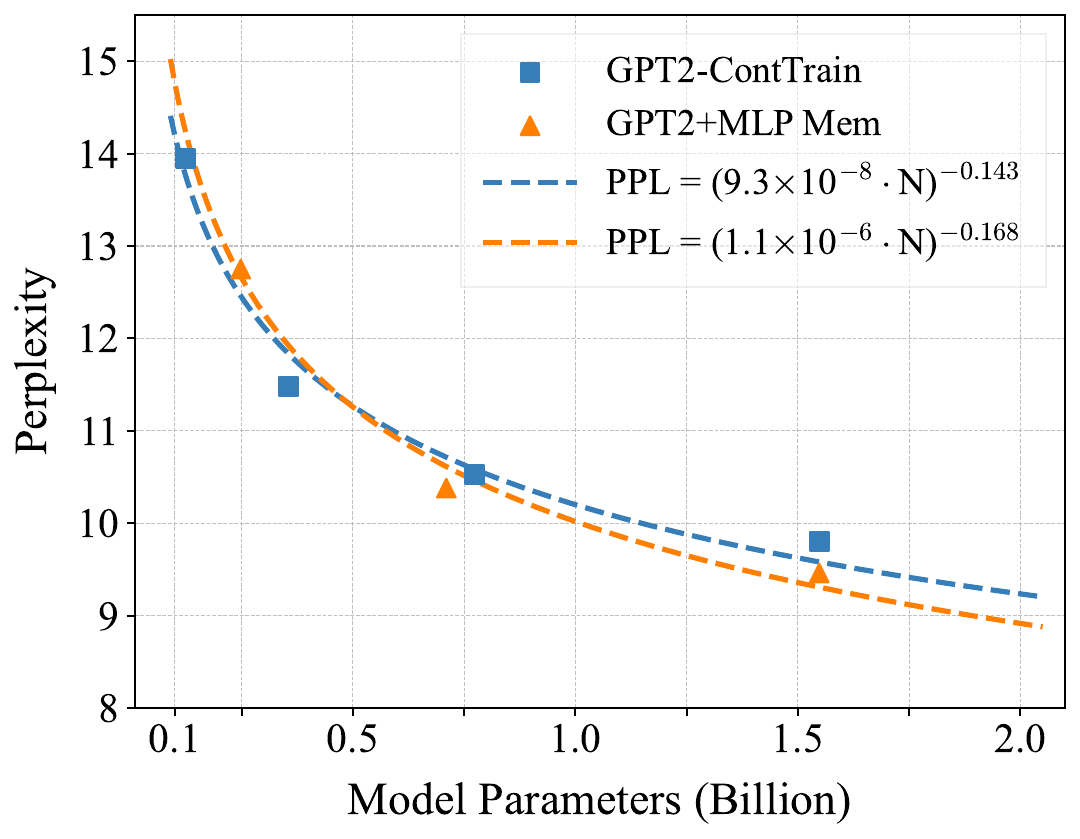}
    \subcaption{Scaling law on WikiText-103}
    \label{fig:scaling_law_a}
  \end{subfigure}
  \hfill 
  \hspace{0.015\textwidth}
  \begin{subfigure}[b]{0.32\textwidth}
    \centering
    \includegraphics[width=\textwidth]{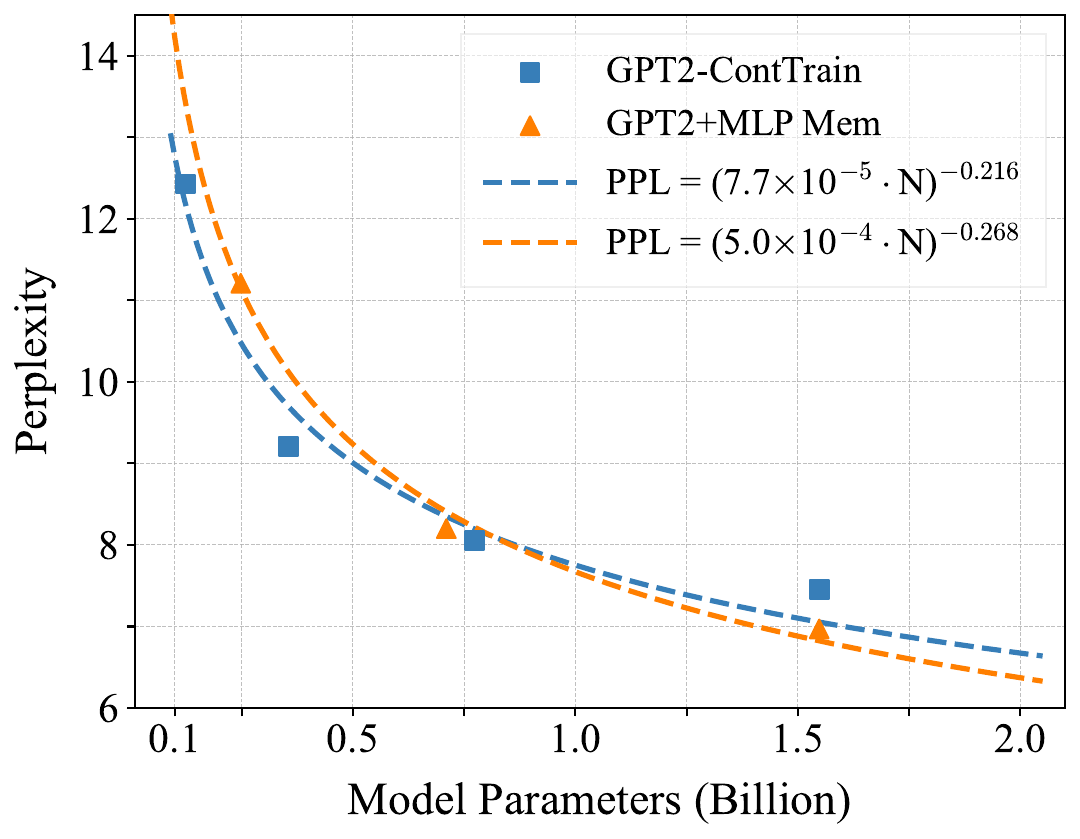}
    \subcaption{Scaling law on Web}
    \label{fig:scaling_law_b}
  \end{subfigure}
  \hfill
  \begin{subfigure}[b]{0.32\textwidth}
    \centering
    \includegraphics[width=\textwidth]{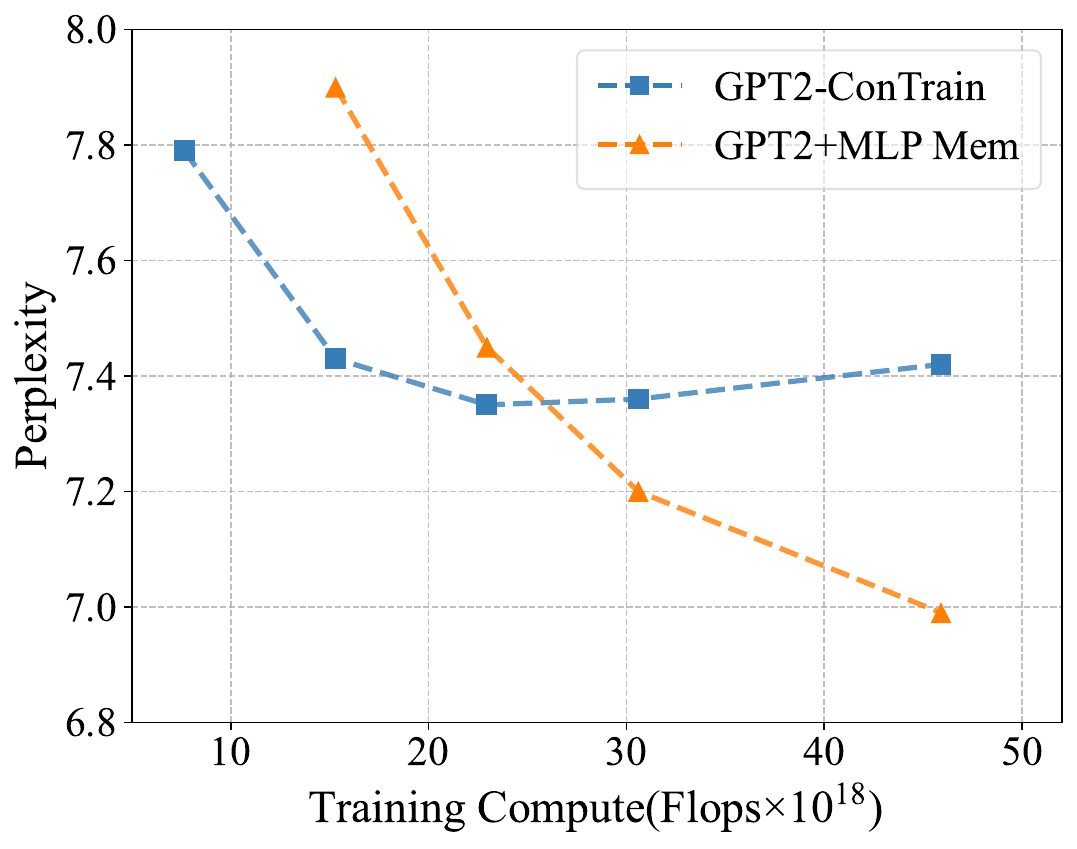}
    \subcaption{Training compute Scaling Law}
    \label{fig:scaling_law_c}
  \end{subfigure}

  \caption{Power-law scaling behavior with model size $N$ and training compute $C$. (a) Scaling results compare the continued training of GPT2 (GPT2-ConTrain) with our overall model architecture (GPT2+MLP Mem) under fixed compute. Our fitted curve shows a 17.5\% exponent improvement on WikiText-103. (b) On the larger Web dataset, our architecture exhibits stronger scaling gains from increased data size, with an exponent improvement of 24.1\%. (c) At the GPT2-xl scale, our architecture continues to benefit from additional training on the Web dataset without overfitting.
  }
  \label{fig:scaling_law}
\end{figure}

\paragraph{Scaling law with model parameters $N$} Following~\cite{scaling_law_openai}, we model perplexity scaling as $PPL=(\beta\cdot N)^{\gamma}$. Under fixed compute, we compare our architecture to continued GPT-2 training on WikiText-103 and Web datasets in terms of test perplexity scaling with model size $N$. Results in Figure \ref{fig:scaling_law} show our architecture demonstrates a steeper scaling curve than the decoder-only model, indicating improved scaling efficiency. The power-law scaling laws on WikiText-103 can be expressed as:
\begin{equation}
    \begin{aligned}
        PPL_d = (9.3\cdot 10^{-8}N)^{-0.143} \quad and \quad PPL_m =(1.1\cdot 10^{-6}N)^{-0.168}
    \end{aligned}
\end{equation}

where $PPL_d$ and $PPL_m$ denote the test perplexity of the decoder model and our architecture, respectively. The corresponding power-law scaling laws on the Web dataset are as follows:
\begin{equation}
    \begin{aligned}
        PPL_d = (7.7 \cdot10^{-5} N)^{-0.216} \quad and \quad PPL_m =(5.0\cdot 10^{-4}N)^{-0.268}
    \end{aligned}
\end{equation}

These results highlight the superior scaling efficiency of our overall model architecture compared to the standard decoder-only baseline, on both WikiText-103 and the Web dataset.

\paragraph{Scaling law with training compute $C$} We further examine how model performance scales with training compute $C$ while keeping model size fixed. At the GPT2-xl scale, we conduct experiments on the Web dataset, measuring test perplexity after varying amounts of training flops. As illustrated in Figure \ref{fig:scaling_law} (c), our overall model architecture achieves significantly lower perplexity with increasing training compute, with no signs of overfitting. This suggests that the retriever imitation pretraining task is more challenging and continues to benefit from additional compute.

\begin{table*}[t]
\centering
\caption{Question answering performance across five benchmarks. Positive gains are shown in \textcolor{green!60!black}{green} and negative changes in \textcolor{red!80!}{red}. \textcolor{green!60!black}{Percentage} in parentheses denotes the relative improvement over the base model. All methods use the same Wikipedia-2021 corpus for training or retrieval.}
\resizebox{\textwidth}{!}{
\begin{tabular}{lllllll}
\toprule
\multirow{2}{*}{\centering Methods} & \multicolumn{3}{c}{Open-Domain QA} & \multicolumn{1}{c}{Long-form QA} & \multicolumn{1}{c}{Multihop QA} & \multirow{2}{*}{Average} \\ 
\cmidrule(lr){2-4} \cmidrule(lr){5-5} \cmidrule(lr){6-6}
 & NQ & WebQA & TriviaQA & TruthfulQA & \multicolumn{1}{c}{HotpotQA} & \\
\midrule
Llama2-7B        &23.18 &32.09 &56.91 &29.16 &22.72 & 32.81 \\

\rowcolor{gray!8}\multicolumn{7}{c}{\textit{Non-parametric methods}} \\

+RAG        &14.60\scriptsize \textcolor{red!80!}{$-$8.58} & 36.71\scriptsize \textcolor{green!60!black}{$+$4.62} & 62.20\scriptsize \textcolor{green!60!black}{$+$5.29} & 31.59\scriptsize \textcolor{green!60!black}{$+$2.43} & 19.60\scriptsize \textcolor{red!80!}{$-$3.12} & 32.94(\textcolor{green!60!black}{$+$0.4\%}) \\

+$k$NN-LM        &23.16\scriptsize \textcolor{red!80!}{$-$0.02} & 33.46\scriptsize \textcolor{green!60!black}{$+$1.37} & 57.31\scriptsize \textcolor{green!60!black}{$+$0.40} & 29.22\scriptsize \textcolor{green!60!black}{$+$0.06} & 22.66\scriptsize \textcolor{red!80!}{$-$0.06} & 33.16(\textcolor{green!60!black}{$+$1.1\%}) \\

\rowcolor{gray!8}\multicolumn{7}{c}{\textit{Parametric methods}} \\

+CPT       &12.90\scriptsize \textcolor{red!80!}{$-$10.28} & 31.55\scriptsize \textcolor{red!80!}{$-$0.54} & 58.81\scriptsize \textcolor{green!60!black}{$+$1.90} & 29.56\scriptsize \textcolor{green!60!black}{$+$0.40} & 15.49\scriptsize \textcolor{red!80!}{$-$7.23} & 29.66(\textcolor{red!80!}{$-$9.6\%}) \\

+LoRA       &17.88\scriptsize \textcolor{red!80!}{$-$5.30} &35.19\scriptsize \textcolor{green!60!black}{$+$3.10} & 58.14\scriptsize \textcolor{green!60!black}{$+$1.23} & 28.33\scriptsize \textcolor{red!80!}{$-$0.83} & 17.18\scriptsize \textcolor{red!80!}{$-$5.54} & 31.34(\textcolor{red!80!}{$-$4.5\%}) \\
\rowcolor{SkyBlue!20}+MLP Mem      &27.04\scriptsize \textcolor{green!60!black}{$+$3.86} & 36.61\scriptsize \textcolor{green!60!black}{$+$4.52} & 57.50\scriptsize \textcolor{green!60!black}{$+$0.59} & 30.04\scriptsize \textcolor{green!60!black}{$+$0.88} & 25.69\scriptsize \textcolor{green!60!black}{$+$2.97}& \textbf{35.38}(\textcolor{green!60!black}{$+$7.8\%}) \\

\midrule
Mistral-7B-v0.3        &20.63 &29.28 &57.65 &32.09 &20.96 & 32.12 \\

\rowcolor{gray!8} \multicolumn{7}{c}{\textit{Non-parametric methods}} \\

+RAG        &22.56\scriptsize\textcolor{green!60!black}{$+$1.93} & 24.90\scriptsize \textcolor{red!80!}{$-$4.38} & 54.21\scriptsize \textcolor{red!80!}{$-$3.44} & 35.47\scriptsize\textcolor{green!60!black}{$+$3.38} & 29.77\scriptsize\textcolor{green!60!black}{$+$8.81} & 33.38(\textcolor{green!60!black}{$+$3.9\%}) \\

+$k$NN-LM        &21.05\scriptsize\textcolor{green!60!black}{$+$0.42} & 30.51\scriptsize \textcolor{green!60!black}{$+$1.23} & 57.77\scriptsize \textcolor{green!60!black}{$+$0.12} & 32.33\scriptsize\textcolor{green!60!black}{$+$0.24} & 21.20\scriptsize\textcolor{green!60!black}{$+$0.24} & 32.57(\textcolor{green!60!black}{$+$1.4\%}) \\

\rowcolor{gray!8} \multicolumn{7}{c}{\textit{Parametric methods}} \\

+CPT       &12.16\scriptsize \textcolor{red!80!}{$-$8.47} & 34.06\scriptsize\textcolor{green!60!black}{$+$4.78} & 61.21\scriptsize\textcolor{green!60!black}{$+$3.56} & 29.18\scriptsize \textcolor{red!80!}{$-$2.91} & 16.04\scriptsize \textcolor{red!80!}{$-$4.92} & 30.53(\textcolor{red!80!}{$-$5.0\%}) \\

+LoRA       &18.17\scriptsize \textcolor{red!80!}{$-$2.46} &34.50\scriptsize\textcolor{green!60!black}{$+$5.22} & 61.60\scriptsize\textcolor{green!60!black}{$+$3.95} & 30.91\scriptsize \textcolor{red!80!}{$-$1.18} & 16.23\scriptsize \textcolor{red!80!}{$-$4.73} & 32.28(\textcolor{green!60!black}{$+$0.5\%}) \\

\rowcolor{SkyBlue!20}+MLP Mem      &25.20\scriptsize \textcolor{green!60!black}{$+$4.57} & 37.45\scriptsize \textcolor{green!60!black}{$+$8.17} & 60.99\scriptsize \textcolor{green!60!black}{$+$3.34} & 32.54\scriptsize \textcolor{green!60!black}{$+$0.45} & 24.14\scriptsize \textcolor{green!60!black}{$+$3.18} & \textbf{36.06}(\textcolor{green!60!black}{+12.3\%}) \\

\bottomrule
\end{tabular}}
\label{tab:qa_downstream}
\end{table*}

\subsection{Question Answering Performance}
\label{sec.5.2}
We evaluate MLP Memory on five diverse QA benchmarks: Natural Questions (NQ)~\citep{NQ}, WebQA~\citep{Webqa}, TriviaQA~\citep{Triviaqa}, TruthfulQA~\citep{truthfulqa}, and HotpotQA~\citep{Hotpotqa}, comparing against CPT, LoRA, and RAG. As shown in Table \ref{tab:qa_downstream}, Mistral-7B-v0.3 with MLP Memory achieves an average relative improvement of 12.3\% over the baseline across five benchmarks, with particularly striking improvements on NQ (25.20\% vs. baseline 20.63\%) and WebQA (37.45\% vs. baseline 29.28\%). While CPT and LoRA suffer significant degradation across all tasks—likely due to catastrophic forgetting during domain-specific training—MLP Memory maintains or improves performance by learning to emulate retrieval behavior without modifying the base model's parameters. Notably, our approach outperforms both RAG and $k$NN-LM, even though they leverage the same Wikipedia-2021 corpus for retrieval at inference time, suggesting that our parametric compression of retrieval patterns captures richer contextual relationships than explicit document retrieval. The consistent gains across both factoid QA (NQ, TriviaQA) and multi-hop reasoning (HotpotQA) demonstrate that MLP Memory effectively bridges the gap between parametric and non-parametric memory systems.








\begin{table*}[t]
\centering
\caption{Performance on nine general NLP tasks spanning sentiment classification, textual entailment, and topic classification.  \textcolor{green!60!black}{↑} indicate improvement over the Mistral-7B-v0.3 baseline, while  \textcolor{red!80!}{↓} indicate decreased performance.}
\resizebox{\textwidth}{!}{
\begin{tabular}{lcccccccccc}
\toprule
\multirow{2}{*}{Methods} & \multicolumn{5}{c}{Sentiment Classification} & \multicolumn{2}{c}{Textual.} & \multicolumn{2}{c}{Topic.} & \multirow{2}{*}{Average} \\
\cmidrule(lr){2-6} \cmidrule(lr){7-8} \cmidrule(lr){9-10}
 & SST2 & MR & CR & RT & HYP & CB & RTE & AGN & Yahoo & \\ \midrule
Mistral-7B-v0.3 &81.21 &75.35 &62.30 &74.95 &55.42 & 69.64 & 59.57 & 75.95 & 56.36 & 67.86 \\
\rowcolor{gray!8}\multicolumn{11}{c}{\textit{Non-parametric methods}} \\
+RAG & 87.20\textcolor{green!60!black}{↑} & 83.70\textcolor{green!60!black}{↑} & 71.55\textcolor{green!60!black}{↑} & 82.36\textcolor{green!60!black}{↑} & 54.65\textcolor{red!80!}{↓} & 57.14\textcolor{red!80!}{↓} & 66.43\textcolor{green!60!black}{↑} & 75.64\textcolor{red!80!}{↓} & 58.43\textcolor{green!60!black}{↑}
& 70.79\textcolor{green!60!black}{↑} \\
+$k$NN-LM  & 82.15\textcolor{green!60!black}{↑} & 76.85\textcolor{green!60!black}{↑} & 61.70\textcolor{red!80!}{↓} & 74.95 & 56.78\textcolor{green!60!black}{↑} & 71.42\textcolor{green!60!black}{↑} & 60.28\textcolor{green!60!black}{↑} & 76.13\textcolor{green!60!black}{↑} & 56.26\textcolor{red!80!}{↓} & 68.50\textcolor{green!60!black}{↑} \\
\rowcolor{gray!8}\multicolumn{11}{c}{\textit{Parametric methods}} \\
+CPT &87.09\textcolor{green!60!black}{↑} &82.85\textcolor{green!60!black}{↑} & 82.60\textcolor{green!60!black}{↑} & 77.48\textcolor{green!60!black}{↑} & 60.65\textcolor{green!60!black}{↑} & 57.14\textcolor{red!80!}{↓} & 52.71\textcolor{red!80!}{↓} & 83.10\textcolor{green!60!black}{↑} & 51.56\textcolor{red!80!}{↓}
& 70.58\textcolor{green!60!black}{↑} \\
+LoRA & 86.54\textcolor{green!60!black}{↑} &83.20\textcolor{green!60!black}{↑} &75.10\textcolor{green!60!black}{↑} & 79.83\textcolor{green!60!black}{↑} & 55.42 & 51.78\textcolor{red!80!}{↓} & 56.31\textcolor{red!80!}{↓} & 65.46\textcolor{red!80!}{↓} & 57.30\textcolor{green!60!black}{↑}
& 67.88\textcolor{green!60!black}{↑} \\
\rowcolor{SkyBlue!20}
+MLP Mem & 83.19\textcolor{green!60!black}{↑} & 79.90\textcolor{green!60!black}{↑} & 75.95\textcolor{green!60!black}{↑} & 75.42\textcolor{green!60!black}{↑} & 64.15\textcolor{green!60!black}{↑} & 76.79\textcolor{green!60!black}{↑} & 64.62\textcolor{green!60!black}{↑} & 80.28\textcolor{green!60!black}{↑} & 57.33\textcolor{green!60!black}{↑}
& \textbf{73.07}\textcolor{green!60!black}{↑} \\
\bottomrule
\end{tabular}}
\label{tab:memory-intensive}
\end{table*}

\subsection{General NLP Task Performance}
\label{general_nlp_task}
To ensure MLP Memory doesn't compromise fundamental language understanding, we evaluate on nine standard NLP tasks spanning sentiment classification (SST-2~\citep{SST2}, MR~\citep{MR}, CR~\citep{CR}, RT~\citep{RT}, HYP~\citep{HYP}), textual entailment (CB~\citep{CB}, RTE~\citep{RTE}), and topic classification (AGNews~\citep{AGN}, Yahoo~\citep{Yahoo}). Table \ref{tab:memory-intensive} reveals that MLP Memory achieves comprehensive improvements across all general tasks, achieving the highest average score compared to all baselines. The improvements are particularly pronounced on reasoning-intensive tasks like RTE (64.62\% vs. baseline 59.57\%) and CB (76.79\% vs. baseline 69.64\%), suggesting that the retrieval patterns learned by MLP Memory provide useful inductive biases even for tasks that don't explicitly require factual knowledge retrieval. In contrast, CPT and LoRA show mixed results with improvements on some tasks but degradation on others. The robust performance across this diverse task suite demonstrates that MLP Memory's external parametric memory complements rather than interferes with the base model's learned representations.

\subsection{Hallucination Reduction}
\label{sec.5.4}

\begin{table*}[t]
\centering
\caption{Performance on HaluEval benchmark across question answering, dialogue, and summarization tasks. Results show accuracy (\%). RAG is not evaluated on summarization as this task requires only the source document.}

\begin{tabular}{llll}
\toprule
\multirow{1}{*}{\centering} & \multicolumn{1}{c}{Dialogue} & \multicolumn{1}{c}{QA} & \multicolumn{1}{c}{Summarization} \\ 
\midrule
Mistral-7B-v0.3        &57.18 &53.99 &50.27 \\

+CPT & 51.68\scriptsize\textcolor{red!80!}{$-$5.50} & 46.49\scriptsize\textcolor{red!80!}{$-$7.50} & 47.39\scriptsize\textcolor{red!80!}{$-$2.88} \\

+LoRA &55.51\scriptsize\textcolor{red!80!}{$-$1.67}&50.02\scriptsize\textcolor{red!80!}{$-$3.97}&50.38\scriptsize\textcolor{green!60!black}{$+$0.11} \\

+RAG        &59.06\scriptsize\textcolor{green!60!black}{$+$1.88} & 65.09\scriptsize\textcolor{green!60!black}{$+$11.10} & - \\

\rowcolor{SkyBlue!20}+MLP Mem      &66.86\scriptsize\textcolor{green!60!black}{$+$9.68}& 64.07\scriptsize\textcolor{green!60!black}{$+$10.08} & 52.41\scriptsize\textcolor{green!60!black}{$+$2.14} \\ 
\bottomrule
\end{tabular}
\label{tab:halucination-downstream}
\end{table*}

We assess MLP Memory's ability to reduce hallucinations using HaluEval~\citep{halueval} across three generation tasks: dialogue, question answering, and summarization, where models must identify factual inconsistencies in generated content. As shown in Table \ref{tab:halucination-downstream}, parametric methods (CPT and LoRA) severely degrade performance, confirming the risks of weight modification. MLP Memory consistently improves hallucination detection across all three domains, with gains of 9.68, 10.08, and 2.14 points respectively. These substantial improvements indicate that the retrieval patterns encoded in MLP Memory significantly help the model better distinguish factual from hallucinated content. The effectiveness across diverse generation contexts—from free-form dialogue to constrained summarization—suggests that MLP Memory's learned retrieval behavior provides a general mechanism for grounding language generation in factual knowledge. This hallucination reduction, combined with strong QA performance and enhanced general capabilities, validates our core hypothesis that decoupling memorization from reasoning through retriever-pretrained external memory can enhance factual accuracy without the typical trade-offs of parametric or non-parametric approaches.

\begin{figure}[t]
\centering
\includegraphics[width=\textwidth]{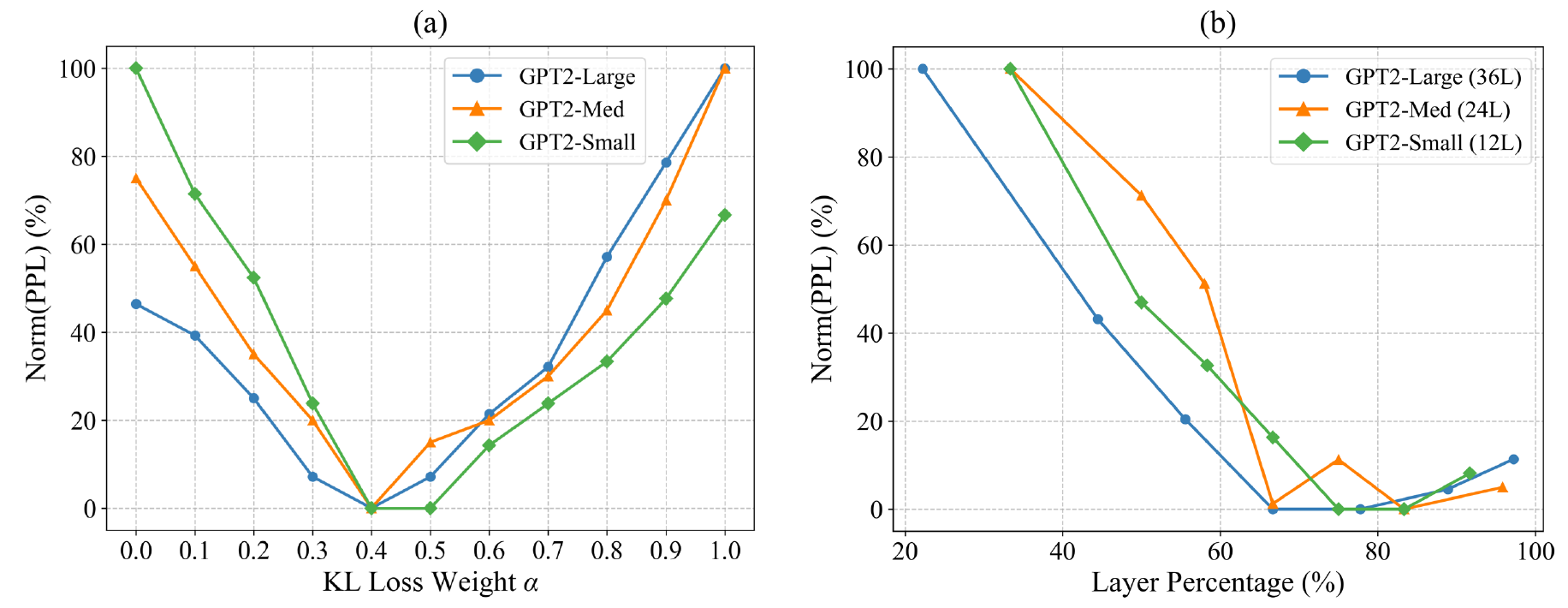}
\caption{(a) Impact of KL loss weight $\alpha$ on retriever imitation. Lower PPL (min-max normalized for clarity) indicates better performance, with optimal balance at $\alpha = 0.4$. (b) Impact of input layer depth on MLP Memory performance across model sizes. Layer percentage denotes depth in the decoder stack (e.g., 70\% corresponds to layer 25 in GPT2-large).}
\label{fig:ablation_study}
\end{figure}

\subsection{Ablation study}

\label{ablation_study}
\textbf{Ablation Setup\quad}
We conduct ablation experiments across three GPT2~\citep{gpt2} scales: small (12 layers), medium (24 layers), and large (36 layers), paired with corresponding MLP Memory modules of 117M, 345M, and 774M parameters respectively. All experiments are evaluated on WikiText-103~\citep{Wikitext103} to investigate loss weighting and optimal layer selection.

\textbf{Impact of Loss Weighting\quad} We examine how balancing KL and CE losses affects retriever imitation by varying $\alpha$ from 0.0 to 1.0. As Figure~\ref{fig:ablation_study}(a) shows, extreme values produce suboptimal results—low values prevent the MLP memory from learning from the $k$NN distribution, while high values cause overfitting to the language modeling objective. The optimal balance occurs at $\alpha=0.4$, indicating both objectives are necessary. KL divergence leverages the information-rich kNN distribution, enabling more effective generalization, while CE loss provides essential token-level prediction accuracy. This balanced approach prevents overfitting while maintaining predictive power.

\textbf{Which Layer Provides the Best Representation for MLP Memory?\quad} While $k$NN-LM performs best using the input to the final feedforward layer as the retrieval key, our MLP Memory consistently achieves optimal performance at around 70\% of network depth, regardless of model scale. Our finding aligns with Memorizing Transformers~\citep{MemoryTransformer}, which also selected around 75\% depth for optimal retrieval performance. We evaluate GPT2-small (12 layers), GPT2-medium (24 layers), and GPT2-large (36 layers), attaching the MLP Memory to various transformer blocks. As shown in Figure~\ref{fig:ablation_study}(b), the x-axis indicates relative depth (20\%–100\%), and the y-axis shows min-max normalized perplexity (0\% = best, 100\% = worst). This consistent pattern across all model sizes contrasts with the $k$NN-LM convention of using final-layer representations.

\section{Related Work}

\textbf{Retrieval-Augmented Generation\quad}
RAG~\citep{RAG1, RAG2, RAG3} mitigates hallucinations by grounding generation in external knowledge. Despite improving factual accuracy, RAG faces limitations: retrieval latency, coarse granularity, and limited LLM integration~\citep{ARL2}. Recent work~\citep{PRAG} explores enhanced retrieval with LLM priors. Our approach proposes a parametric memory mimicking non-parametric retrieval, eliminating explicit document retrieval while preserving knowledge augmentation.

\textbf{Memory-Augmented Language Models\quad}
Various architectures explored memory augmentation, from Memory Networks~\citep{MemoryNetworks} with explicit read-write components to Memory Transformers~\citep{MemoryTransformer} with extended attention. LongMem~\citep{LongMem} and MemoRAG~\citep{MemoRAG} introduced decoupled architectures for long-term history storage. While these focus on context extension, our MLP memory expands across the entire pre-training corpus, enabling long-term generalizable knowledge storage.

\textbf{MLP Architectures\quad}
All-MLP architectures emerged as transformer alternatives, with gMLP~\citep{gMLP} matching transformer performance and sparse MLPs~\citep{sparse-all-MLP} showing superior training efficiency. Studies~\citep{FFN-KV} identified FFN layers as key-value memories in LLMs. Inspired by this, we propose pretraining an all-MLP memory as a non-parametric retriever, leveraging MLPs' memorization capabilities for a compact, differentiable knowledge store.

\section{Conclusion}

In this paper, we introduced MLP Memory, a novel approach for enhancing language models by learning to internalize retrieval patterns. By pretraining a lightweight MLP module to imitate kNN retriever behavior on the entire pretraining corpus, MLP Memory captures the benefits of retrieval-augmented generation in a fully parametric form, without requiring explicit document access.

The key advantage of MLP Memory lies in its efficiency and effectiveness, as our architecture exhibits stronger scaling behavior than decoder-only models. On downstream tasks, our approach achieves 12.3\% relative improvement on question-answering benchmarks, 5.2 points gain on general NLP tasks, and up to 10 points reduction in hallucinations—while delivering 2.5$\times$ faster inference than RAG and maintaining constant speed regardless of corpus size. Unlike parametric fine-tuning that risks catastrophic forgetting or non-parametric RAG that suffers from high latency, MLP Memory enhances model capabilities without these typical trade-offs.

MLP Memory introduces a new paradigm for language model enhancement that fundamentally reimagines how models access and utilize knowledge. By parametrically encoding retrieval behavior through a pretrained memory component, our approach creates a more efficient, accurate, and scalable framework that bridges the gap between parametric and non-parametric methods.
\section*{Acknowledgments}
This work is sponsored by the National Natural Science Foundation of China (NSFC) grant (No. 62576211) and the National Key Research and Development Program of China (No. 2023ZD0121402). It is also the result of a collaborative project on novel language model architectures between Shanghai Jiao Tong University (SJTU) and the Shanghai Artificial Intelligence Laboratory. The computational resources required for pretraining the models were provided by the Shanghai AI Lab. This work is also supported by the Specialized Program on Fundamental Research from Science and Technology Commission of Shanghai Municipality (No. 2025SHZDZX025G09).


\bibliography{references}

@article{gpt2,
  title={Language models are unsupervised multitask learners},
  author={Radford, Alec and Wu, Jeffrey and Child, Rewon and Luan, David and Amodei, Dario and Sutskever, Ilya and others},
  journal={OpenAI blog},
  volume={1},
  number={8}, 
  pages={9},
  year={2019}
}

@misc{gpt3,
      title={Language Models are Few-Shot Learners}, 
      author={Tom B. Brown and Benjamin Mann and Nick Ryder and Melanie Subbiah and Jared Kaplan and Prafulla Dhariwal and Arvind Neelakantan and Pranav Shyam and Girish Sastry and Amanda Askell and Sandhini Agarwal and Ariel Herbert-Voss and Gretchen Krueger and Tom Henighan and Rewon Child and Aditya Ramesh and Daniel M. Ziegler and Jeffrey Wu and Clemens Winter and Christopher Hesse and Mark Chen and Eric Sigler and Mateusz Litwin and Scott Gray and Benjamin Chess and Jack Clark and Christopher Berner and Sam McCandlish and Alec Radford and Ilya Sutskever and Dario Amodei},
      year={2020},
      eprint={2005.14165},
      archivePrefix={arXiv},
      primaryClass={cs.CL},
      url={https://arxiv.org/abs/2005.14165}, 
}

@misc{gpt4,
      title={GPT-4 Technical Report}, 
      author={OpenAI and Josh Achiam and Steven Adler and Sandhini Agarwal and Lama Ahmad and Ilge Akkaya and Florencia Leoni Aleman and Diogo Almeida and Janko Altenschmidt and Sam Altman and Shyamal Anadkat},
      year={2024},
      eprint={2303.08774},
      archivePrefix={arXiv},
      primaryClass={cs.CL},
      url={https://arxiv.org/abs/2303.08774}, 
}

@article{llama2,
  title={Llama 2: Open foundation and fine-tuned chat models},
  author={Touvron, Hugo and Martin, Louis and Stone, Kevin and Albert, Peter and Almahairi, Amjad and Babaei, Yasmine and Bashlykov, Nikolay and Batra, Soumya and Bhargava, Prajjwal and Bhosale, Shruti and others},
  journal={arXiv preprint arXiv:2307.09288},
  year={2023}
}

@article{llama3,
  title={The llama 3 herd of models},
  author={Grattafiori, Aaron and Dubey, Abhimanyu and Jauhri, Abhinav and Pandey, Abhinav and Kadian, Abhishek and Al-Dahle, Ahmad and Letman, Aiesha and Mathur, Akhil and Schelten, Alan and Vaughan, Alex and others},
  journal={arXiv preprint arXiv:2407.21783},
  year={2024}
}

@misc{qwen2.5,
      title={Qwen2.5 Technical Report}, 
      author={Qwen and : and An Yang and Baosong Yang and Beichen Zhang and Binyuan Hui and Bo Zheng and Bowen Yu and Chengyuan Li and Dayiheng Liu and Fei Huang and Haoran Wei and Huan Lin and Jian Yang and Jianhong Tu and Jianwei Zhang and Jianxin Yang and Jiaxi Yang and Jingren Zhou and Junyang Lin and Kai Dang and Keming Lu and Keqin Bao and Kexin Yang and Le Yu and Mei Li and Mingfeng Xue and Pei Zhang and Qin Zhu and Rui Men and Runji Lin and Tianhao Li and Tianyi Tang and Tingyu Xia and Xingzhang Ren and Xuancheng Ren and Yang Fan and Yang Su and Yichang Zhang and Yu Wan and Yuqiong Liu and Zeyu Cui and Zhenru Zhang and Zihan Qiu},
      year={2025},
      eprint={2412.15115},
      archivePrefix={arXiv},
      primaryClass={cs.CL},
      url={https://arxiv.org/abs/2412.15115}, 
}

@article{deepseekv3,
  title={Deepseek-v3 technical report},
  author={Liu, Aixin and Feng, Bei and Xue, Bing and Wang, Bingxuan and Wu, Bochao and Lu, Chengda and Zhao, Chenggang and Deng, Chengqi and Zhang, Chenyu and Ruan, Chong and others},
  journal={arXiv preprint arXiv:2412.19437},
  year={2024}
}

@article{llava,
  title={Visual instruction tuning},
  author={Liu, Haotian and Li, Chunyuan and Wu, Qingyang and Lee, Yong Jae},
  journal={Advances in neural information processing systems},
  volume={36},
  pages={34892--34916},
  year={2023}
}

@article{gpt-code,
  title={Evaluating large language models trained on code},
  author={Chen, Mark and Tworek, Jerry and Jun, Heewoo and Yuan, Qiming and Pinto, Henrique Ponde De Oliveira and Kaplan, Jared and Edwards, Harri and Burda, Yuri and Joseph, Nicholas and Brockman, Greg and others},
  journal={arXiv preprint arXiv:2107.03374},
  year={2021}
}

@inproceedings{GPT-image,
author = {Chen, Mark and Radford, Alec and Child, Rewon and Wu, Jeff and Jun, Heewoo and Luan, David and Sutskever, Ilya},
title = {Generative pretraining from pixels},
year = {2020},
publisher = {JMLR.org},
booktitle = {Proceedings of the 37th International Conference on Machine Learning},
articleno = {158},
numpages = {13},
series = {ICML'20}
}

@misc{RAG1,
      title={Retrieval-Augmented Generation for Knowledge-Intensive NLP Tasks}, 
      author={Patrick Lewis and Ethan Perez and Aleksandra Piktus and Fabio Petroni and Vladimir Karpukhin and Naman Goyal and Heinrich Küttler and Mike Lewis and Wen-tau Yih and Tim Rocktäschel and Sebastian Riedel and Douwe Kiela},
      year={2021},
      eprint={2005.11401},
      archivePrefix={arXiv},
      primaryClass={cs.CL},
      url={https://arxiv.org/abs/2005.11401}, 
}

@misc{RAG2,
      title={Check Your Facts and Try Again: Improving Large Language Models with External Knowledge and Automated Feedback}, 
      author={Baolin Peng and Michel Galley and Pengcheng He and Hao Cheng and Yujia Xie and Yu Hu and Qiuyuan Huang and Lars Liden and Zhou Yu and Weizhu Chen and Jianfeng Gao},
      year={2023},
      eprint={2302.12813},
      archivePrefix={arXiv},
      primaryClass={cs.CL},
      url={https://arxiv.org/abs/2302.12813}, 
}

@misc{RAG-atlas,
      title={Atlas: Few-shot Learning with Retrieval Augmented Language Models}, 
      author={Gautier Izacard and Patrick Lewis and Maria Lomeli and Lucas Hosseini and Fabio Petroni and Timo Schick and Jane Dwivedi-Yu and Armand Joulin and Sebastian Riedel and Edouard Grave},
      year={2022},
      eprint={2208.03299},
      archivePrefix={arXiv},
      primaryClass={cs.CL},
      url={https://arxiv.org/abs/2208.03299}, 
}

@misc{RAG3,
      title={Precise Zero-Shot Dense Retrieval without Relevance Labels}, 
      author={Luyu Gao and Xueguang Ma and Jimmy Lin and Jamie Callan},
      year={2022},
      eprint={2212.10496},
      archivePrefix={arXiv},
      primaryClass={cs.IR},
      url={https://arxiv.org/abs/2212.10496}, 
}

@incollection{MLP,
  title={Multilayer perceptron (MLP)},
  author={Taud, Hind and Mas, Jean-Franccois},
  booktitle={Geomatic approaches for modeling land change scenarios},
  pages={451--455},
  year={2017},
  publisher={Springer}
}

@misc{MemoryNetworks,
      title={Memory Networks}, 
      author={Jason Weston and Sumit Chopra and Antoine Bordes},
      year={2015},
      eprint={1410.3916},
      archivePrefix={arXiv},
      primaryClass={cs.AI},
      url={https://arxiv.org/abs/1410.3916}, 
}

@misc{MemoryTransformer,
      title={Memory Transformer}, 
      author={Mikhail S. Burtsev and Yuri Kuratov and Anton Peganov and Grigory V. Sapunov},
      year={2021},
      eprint={2006.11527},
      archivePrefix={arXiv},
      primaryClass={cs.CL},
      url={https://arxiv.org/abs/2006.11527}, 
}

@misc{AutoCompressor,
      title={Adapting Language Models to Compress Contexts}, 
      author={Alexis Chevalier and Alexander Wettig and Anirudh Ajith and Danqi Chen},
      year={2023},
      eprint={2305.14788},
      archivePrefix={arXiv},
      primaryClass={cs.CL},
      url={https://arxiv.org/abs/2305.14788}, 
}

@misc{Wikitext103,
      title={Pointer Sentinel Mixture Models}, 
      author={Stephen Merity and Caiming Xiong and James Bradbury and Richard Socher},
      year={2016},
      eprint={1609.07843},
      archivePrefix={arXiv},
      primaryClass={cs.CL},
      url={https://arxiv.org/abs/1609.07843}, 
}

@article{Split-Brain,
  title={Forty-five years of split-brain research and still going strong},
  author={Gazzaniga, Michael S},
  journal={Nature Reviews Neuroscience},
  volume={6},
  number={8},
  pages={653--659},
  year={2005},
  publisher={Nature Publishing Group}
}

@article{hippocampus,
  title={The hippocampus and behavior.},
  author={Douglas, Robert J},
  journal={Psychological bulletin},
  volume={67},
  number={6},
  pages={416},
  year={1967},
  publisher={American Psychological Association}
}

@book{EthicalBrain,
  title={The ethical brain.},
  author={Gazzaniga, Michael S},
  year={2005},
  publisher={Dana press}
}

@misc{knnlm,
      title={Generalization through Memorization: Nearest Neighbor Language Models}, 
      author={Urvashi Khandelwal and Omer Levy and Dan Jurafsky and Luke Zettlemoyer and Mike Lewis},
      year={2020},
      eprint={1911.00172},
      archivePrefix={arXiv},
      primaryClass={cs.CL},
      url={https://arxiv.org/abs/1911.00172}, 
}

@article{KL,
  title={R{\'e}nyi divergence and Kullback-Leibler divergence},
  author={Van Erven, Tim and Harremos, Peter},
  journal={IEEE Transactions on Information Theory},
  volume={60},
  number={7},
  pages={3797--3820},
  year={2014},
  publisher={IEEE}
}

@article{CEloss,
  title={Generalized cross entropy loss for training deep neural networks with noisy labels},
  author={Zhang, Zhilu and Sabuncu, Mert},
  journal={Advances in neural information processing systems},
  volume={31},
  year={2018}
}

@misc{sparse-all-MLP,
      title={Efficient Language Modeling with Sparse all-MLP}, 
      author={Ping Yu and Mikel Artetxe and Myle Ott and Sam Shleifer and Hongyu Gong and Ves Stoyanov and Xian Li},
      year={2022},
      eprint={2203.06850},
      archivePrefix={arXiv},
      primaryClass={cs.CL},
      url={https://arxiv.org/abs/2203.06850}, 
}

@misc{gMLP,
      title={Pay Attention to MLPs}, 
      author={Hanxiao Liu and Zihang Dai and David R. So and Quoc V. Le},
      year={2021},
      eprint={2105.08050},
      archivePrefix={arXiv},
      primaryClass={cs.LG},
      url={https://arxiv.org/abs/2105.08050}, 
}

@misc{Efficient-knnlm,
      title={Efficient Nearest Neighbor Language Models}, 
      author={Junxian He and Graham Neubig and Taylor Berg-Kirkpatrick},
      year={2021},
      eprint={2109.04212},
      archivePrefix={arXiv},
      primaryClass={cs.CL},
      url={https://arxiv.org/abs/2109.04212}, 
}

@article{knnlm-limits,
  title={Great Memory, Shallow Reasoning: Limits of $ k $ NN-LMs},
  author={Geng, Shangyi and Zhao, Wenting and Rush, Alexander M},
  journal={arXiv preprint arXiv:2408.11815},
  year={2024}
}

@misc{Mistral-7B,
      title={Mistral 7B}, 
      author={Albert Q. Jiang and Alexandre Sablayrolles and Arthur Mensch and Chris Bamford and Devendra Singh Chaplot and Diego de las Casas and Florian Bressand and Gianna Lengyel and Guillaume Lample and Lucile Saulnier and Lélio Renard Lavaud and Marie-Anne Lachaux and Pierre Stock and Teven Le Scao and Thibaut Lavril and Thomas Wang and Timothée Lacroix and William El Sayed},
      year={2023},
      eprint={2310.06825},
      archivePrefix={arXiv},
      primaryClass={cs.CL},
      url={https://arxiv.org/abs/2310.06825}, 
}

@misc{truthfulqa,
      title={TruthfulQA: Measuring How Models Mimic Human Falsehoods}, 
      author={Stephanie Lin and Jacob Hilton and Owain Evans},
      year={2022},
      eprint={2109.07958},
      archivePrefix={arXiv},
      primaryClass={cs.CL},
      url={https://arxiv.org/abs/2109.07958}, 
}

@misc{halueval,
      title={HaluEval: A Large-Scale Hallucination Evaluation Benchmark for Large Language Models}, 
      author={Junyi Li and Xiaoxue Cheng and Wayne Xin Zhao and Jian-Yun Nie and Ji-Rong Wen},
      year={2023},
      eprint={2305.11747},
      archivePrefix={arXiv},
      primaryClass={cs.CL},
      url={https://arxiv.org/abs/2305.11747}, 
}

@inproceedings{SST2,
    title = "Recursive Deep Models for Semantic Compositionality Over a Sentiment Treebank",
    author = "Socher, Richard  and
      Perelygin, Alex  and
      Wu, Jean  and
      Chuang, Jason  and
      Manning, Christopher D.  and
      Ng, Andrew  and
      Potts, Christopher",
    editor = "Yarowsky, David  and
      Baldwin, Timothy  and
      Korhonen, Anna  and
      Livescu, Karen  and
      Bethard, Steven",
    booktitle = "Proceedings of the 2013 Conference on Empirical Methods in Natural Language Processing",
    month = oct,
    year = "2013",
    address = "Seattle, Washington, USA",
    publisher = "Association for Computational Linguistics",
    url = "https://aclanthology.org/D13-1170/",
    pages = "1631--1642"
}

@inproceedings{MR,
author = {Pang, Bo and Lee, Lillian},
title = {Seeing stars: exploiting class relationships for sentiment categorization with respect to rating scales},
year = {2005},
publisher = {Association for Computational Linguistics},
address = {USA},
url = {https://doi.org/10.3115/1219840.1219855},
doi = {10.3115/1219840.1219855},
booktitle = {Proceedings of the 43rd Annual Meeting on Association for Computational Linguistics},
pages = {115–124},
numpages = {10},
location = {Ann Arbor, Michigan},
series = {ACL '05}
}

@inproceedings{CR,
author = {Hu, Minqing and Liu, Bing},
title = {Mining and summarizing customer reviews},
year = {2004},
isbn = {1581138881},
publisher = {Association for Computing Machinery},
address = {New York, NY, USA},
url = {https://doi.org/10.1145/1014052.1014073},
doi = {10.1145/1014052.1014073},
booktitle = {Proceedings of the Tenth ACM SIGKDD International Conference on Knowledge Discovery and Data Mining},
pages = {168–177},
numpages = {10},
location = {Seattle, WA, USA},
series = {KDD '04}
}

@inproceedings{HYP,
    title = "{S}em{E}val-2019 Task 4: Hyperpartisan News Detection",
    author = "Kiesel, Johannes  and
      Mestre, Maria  and
      Shukla, Rishabh  and
      Vincent, Emmanuel  and
      Adineh, Payam  and
      Corney, David  and
      Stein, Benno  and
      Potthast, Martin",
    editor = "May, Jonathan  and
      Shutova, Ekaterina  and
      Herbelot, Aurelie  and
      Zhu, Xiaodan  and
      Apidianaki, Marianna  and
      Mohammad, Saif M.",
    booktitle = "Proceedings of the 13th International Workshop on Semantic Evaluation",
    month = jun,
    year = "2019",
    address = "Minneapolis, Minnesota, USA",
    publisher = "Association for Computational Linguistics",
    url = "https://aclanthology.org/S19-2145/",
    doi = "10.18653/v1/S19-2145",
    pages = "829--839"
}

@inproceedings{CB,
  title={The commitmentbank: Investigating projection in naturally occurring discourse},
  author={De Marneffe, Marie-Catherine and Simons, Mandy and Tonhauser, Judith},
  booktitle={proceedings of Sinn und Bedeutung},
  volume={23},
  number={2},
  pages={107--124},
  year={2019}
}

@InProceedings{RT,
  author =       {Bo Pang and Lillian Lee},
  title =        {Seeing stars: Exploiting class relationships for sentiment
                  categorization with respect to rating scales},
  booktitle =    {Proceedings of the ACL},
  year =         2005
}

@article{RTE,
  title={Recognizing textual entailment: Rational, evaluation and approaches--erratum},
  author={Dagan, Ido and Dolan, Bill and Magnini, Bernardo and Roth, Dan},
  journal={Natural Language Engineering},
  volume={16},
  number={1},
  pages={105--105},
  year={2010},
  publisher={Cambridge University Press}
}

@article{AGN,
  title={Character-level convolutional networks for text classification},
  author={Zhang, Xiang and Zhao, Junbo and LeCun, Yann},
  journal={Advances in neural information processing systems},
  volume={28},
  year={2015}
}

@article{Yahoo,
  title={Character-level convolutional networks for text classification},
  author={Zhang, Xiang and Zhao, Junbo and LeCun, Yann},
  journal={Advances in neural information processing systems},
  volume={28},
  year={2015}
}

@article{scaling_law_openai,
  title={Scaling laws for neural language models},
  author={Kaplan, Jared and McCandlish, Sam and Henighan, Tom and Brown, Tom B and Chess, Benjamin and Child, Rewon and Gray, Scott and Radford, Alec and Wu, Jeffrey and Amodei, Dario},
  journal={arXiv preprint arXiv:2001.08361},
  year={2020}
}

@inproceedings{Webdataset,
    title = "Nearest Neighbor Zero-Shot Inference",
    author = "Shi, Weijia  and
      Michael, Julian  and
      Gururangan, Suchin  and
      Zettlemoyer, Luke",
    editor = "Goldberg, Yoav  and
      Kozareva, Zornitsa  and
      Zhang, Yue",
    booktitle = "Proceedings of the 2022 Conference on Empirical Methods in Natural Language Processing",
    month = dec,
    year = "2022",
    address = "Abu Dhabi, United Arab Emirates",
    publisher = "Association for Computational Linguistics",
    url = "https://aclanthology.org/2022.emnlp-main.214/",
    doi = "10.18653/v1/2022.emnlp-main.214",
    pages = "3254--3265"
}

@inproceedings{AmazonReview,
  title={Ups and downs: Modeling the visual evolution of fashion trends with one-class collaborative filtering},
  author={He, Ruining and McAuley, Julian},
  booktitle={proceedings of the 25th international conference on world wide web},
  pages={507--517},
  year={2016}
}

@misc{CC-NEWs,
  author       = {Hamborg, Felix and
                  Meuschke, Norman and
                  Breitinger, Corinna and
                  Gipp, Bela},
  title        = {news-please: A Generic News Crawler and Extractor},
  month        = mar,
  year         = 2017,
  publisher    = {Zenodo},
  doi          = {10.5281/zenodo.4120316},
  url          = {https://doi.org/10.5281/zenodo.4120316},
}

@inproceedings{IMDB,
    title = "Learning Word Vectors for Sentiment Analysis",
    author = "Maas, Andrew L.  and
      Daly, Raymond E.  and
      Pham, Peter T.  and
      Huang, Dan  and
      Ng, Andrew Y.  and
      Potts, Christopher",
    editor = "Lin, Dekang  and
      Matsumoto, Yuji  and
      Mihalcea, Rada",
    booktitle = "Proceedings of the 49th Annual Meeting of the Association for Computational Linguistics: Human Language Technologies",
    month = jun,
    year = "2011",
    address = "Portland, Oregon, USA",
    publisher = "Association for Computational Linguistics",
    url = "https://aclanthology.org/P11-1015/",
    pages = "142--150"
}

@article{ARL2,
  title={Arl2: Aligning retrievers for black-box large language models via self-guided adaptive relevance labeling},
  author={Zhang, Lingxi and Yu, Yue and Wang, Kuan and Zhang, Chao},
  journal={arXiv preprint arXiv:2402.13542},
  year={2024}
}

@misc{PRAG,
      title={Parametric Retrieval Augmented Generation}, 
      author={Weihang Su and Yichen Tang and Qingyao Ai and Junxi Yan and Changyue Wang and Hongning Wang and Ziyi Ye and Yujia Zhou and Yiqun Liu},
      year={2025},
      eprint={2501.15915},
      archivePrefix={arXiv},
      primaryClass={cs.CL},
      url={https://arxiv.org/abs/2501.15915}, 
}

@article{LongMem,
  title={Augmenting language models with long-term memory},
  author={Wang, Weizhi and Dong, Li and Cheng, Hao and Liu, Xiaodong and Yan, Xifeng and Gao, Jianfeng and Wei, Furu},
  journal={Advances in Neural Information Processing Systems},
  volume={36},
  pages={74530--74543},
  year={2023}
}

@inproceedings{MemoRAG,
  title={MemoRAG: Boosting Long Context Processing with Global Memory-Enhanced Retrieval Augmentation},
  author={Qian, Hongjin and Liu, Zheng and Zhang, Peitian and Mao, Kelong and Lian, Defu and Dou, Zhicheng and Huang, Tiejun},
  booktitle={Proceedings of the ACM on Web Conference 2025},
  pages={2366--2377},
  year={2025}
}

@article{FFN-KV,
  title={Transformer feed-forward layers are key-value memories},
  author={Geva, Mor and Schuster, Roei and Berant, Jonathan and Levy, Omer},
  journal={arXiv preprint arXiv:2012.14913},
  year={2020}
}

@article{LoRA,
  title={Lora: Low-rank adaptation of large language models.},
  author={Hu, Edward J and Shen, Yelong and Wallis, Phillip and Allen-Zhu, Zeyuan and Li, Yuanzhi and Wang, Shean and Wang, Lu and Chen, Weizhu and others},
  journal={ICLR},
  volume={1},
  number={2},
  pages={3},
  year={2022}
}

@article{NQ,
  title={Natural questions: a benchmark for question answering research},
  author={Kwiatkowski, Tom and Palomaki, Jennimaria and Redfield, Olivia and Collins, Michael and Parikh, Ankur and Alberti, Chris and Epstein, Danielle and Polosukhin, Illia and Devlin, Jacob and Lee, Kenton and others},
  journal={Transactions of the Association for Computational Linguistics},
  volume={7},
  pages={453--466},
  year={2019},
  publisher={MIT Press One Rogers Street, Cambridge, MA 02142-1209, USA journals-info~…}
}

@article{Triviaqa,
  title={Triviaqa: A large scale distantly supervised challenge dataset for reading comprehension},
  author={Joshi, Mandar and Choi, Eunsol and Weld, Daniel S and Zettlemoyer, Luke},
  journal={arXiv preprint arXiv:1705.03551},
  year={2017}
}

@inproceedings{Webqa,
  title={Semantic parsing on freebase from question-answer pairs},
  author={Berant, Jonathan and Chou, Andrew and Frostig, Roy and Liang, Percy},
  booktitle={Proceedings of the 2013 conference on empirical methods in natural language processing},
  pages={1533--1544},
  year={2013}
}

@article{Hotpotqa,
  title={HotpotQA: A dataset for diverse, explainable multi-hop question answering},
  author={Yang, Zhilin and Qi, Peng and Zhang, Saizheng and Bengio, Yoshua and Cohen, William W and Salakhutdinov, Ruslan and Manning, Christopher D},
  journal={arXiv preprint arXiv:1809.09600},
  year={2018}
}

@article{xrag,
  title={xrag: Extreme context compression for retrieval-augmented generation with one token},
  author={Cheng, Xin and Wang, Xun and Zhang, Xingxing and Ge, Tao and Chen, Si-Qing and Wei, Furu and Zhang, Huishuai and Zhao, Dongyan},
  journal={Advances in Neural Information Processing Systems},
  volume={37},
  pages={109487--109516},
  year={2024}
}

@article{holtzman2021surface,
  title={Surface form competition: Why the highest probability answer isn't always right},
  author={Holtzman, Ari and West, Peter and Shwartz, Vered and Choi, Yejin and Zettlemoyer, Luke},
  journal={arXiv preprint arXiv:2104.08315},
  year={2021}
}

@inproceedings{shi2022nearest,
  title={Nearest neighbor zero-shot inference},
  author={Shi, Weijia and Michael, Julian and Gururangan, Suchin and Zettlemoyer, Luke},
  booktitle={Proceedings of the 2022 Conference on Empirical Methods in Natural Language Processing},
  pages={3254--3265},
  year={2022}
}

@article{bge,
  title={Bge m3-embedding: Multi-lingual, multi-functionality, multi-granularity text embeddings through self-knowledge distillation},
  author={Chen, Jianlv and Xiao, Shitao and Zhang, Peitian and Luo, Kun and Lian, Defu and Liu, Zheng},
  journal={arXiv preprint arXiv:2402.03216},
  year={2024}
}

\newpage

\appendix

\section{Implementation Details}
\label{appendix:implementation_details}

\paragraph{Datasets}For the general NLP tasks in Section \ref{general_nlp_task}, we utilize a heterogeneous corpus constructed by aggregating several publicly available sources that cover diverse domains relevant to common NLP tasks. Following the methodology from~\citep{knnlm-limits}, this corpus comprises WikiText-103~\cite{Wikitext103} for encyclopedic content, Amazon Reviews~\cite{AmazonReview} for user-generated product feedback, CC-NEWS~\cite{CC-NEWs} for journalistic content, and IMDB~\cite{IMDB} for movie reviews and discussions.

This diverse mixture captures both formal and informal language patterns, spans multiple domains from news articles to consumer opinions, and provides comprehensive coverage of linguistic phenomena encountered in real-world NLP applications. The complete dataset is publicly available at: \url{https://huggingface.co/datasets/wentingzhao/knn-prompt-datastore}.

\paragraph{Evaluation Metrics}
For question answering benchmarks, following ~\cite{xrag}, we evaluate three Open Domain Question Answering datasets and HotpotQA using the Exact Match (EM) metric. For long-form QA evaluation, we employ three complementary metrics: MC1, which measures whether the model assigns the highest likelihood to the most accurate answer; MC2, which sums the normalized probabilities over all correct answers; and MC3, which evaluates whether the model assigns a higher average likelihood to true answers than to false ones. We report the average of these three metrics as the final performance measure for long-form QA tasks. For general NLP tasks, following the methodology from ~\cite{shi2022nearest}, we report results using the domain-conditional PMI scoring rule~\cite{holtzman2021surface}. For hallucination reduction evaluation, we use accuracy as the primary metric to assess the model's ability to generate factually correct responses.

\paragraph{Hyperparameters}

In Table~\ref{table:detail_mlp_training}, we list the hyperparameters used for training the 1B MLP Memory model (excluding embedding parameters).

\begin{table*}[ht!]
    \centering
    \caption{Hyperparameters for training the 1B MLP Memory model.}
    \begin{tabular}{cc}
        \toprule
        \textbf{Hyperparameter} & \textbf{Assignment}  \\
        \midrule
        optimizer &  AdamW \\
        learning rate & 4e-4 \\
        lr scheduler type & linear \\
        warmup ratio & 0.03 \\
        weight decay & 0.0 \\
        epochs & 5 \\
        flash attention & False \\
        batch size & 4 \\
        gradient accumulation steps & 4 \\
        num GPUs & 32 \\
        max train samples & 33,000,000 \\
        \bottomrule
    \end{tabular}
    \label{table:detail_mlp_training}
\end{table*}

\section{Sensitivity to Interpolation Weight $\lambda$}
\label{sensitivity_interpolation}
We conducted a comprehensive analysis of our method's sensitivity to the interpolation weight $\lambda$ on the HaluEval benchmark using Mistral-7B-v0.3. Table~\ref{tab:sensitivity_to_lambda} presents the results across three tasks: Dialogue, QA, and Summarization, with $\lambda$ values ranging from 0.1 to 0.9.

Our findings demonstrate that the method exhibits robust performance across a wide range of $\lambda$ values, with optimal performance generally observed in the 0.35-0.55 range. Specifically, the Dialogue task achieves its best performance at $\lambda=0.45$ (64.07\%), QA peaks at $\lambda=0.55$ (66.86\%), and Summarization reaches its maximum at $\lambda=0.35$ (52.41\%). Notably, all three tasks show consistent improvements over the baseline Mistral-7B-v0.3 model across the optimal range, with QA showing the most substantial gains (up to 10.08 points improvement).

The performance remains relatively stable within the 0.3-0.6 range, with only gradual degradation outside this interval. When $\lambda$ approaches extreme values (e.g., 0.9), performance deteriorates significantly, particularly for Dialogue and Summarization tasks, though still maintaining improvements over the baseline in the QA task.

These results confirm that our method is not overly sensitive to the specific choice of $\lambda$ within a reasonable range, making it practical for deployment without extensive hyperparameter tuning. The consistent improvements across different $\lambda$ values and tasks validate the robustness of our approach.

\begin{table*}[h]
\centering
\caption{Performance sensitivity analysis of interpolation weight $\lambda$ on HaluEval benchmark using Mistral-7B-v0.3. Results are reported as accuracy (\%) across three tasks: Dialogue, QA, and Summarization. The first row shows baseline Mistral-7B-v0.3 performance without memory augmentation. Bold values indicate the best performance for each task.}

\begin{tabular}{lccc}
\toprule
\multirow{1}{*}{\centering $\lambda$} & \multicolumn{1}{c}{Dialogue} & \multicolumn{1}{c}{QA} & \multicolumn{1}{c}{Summarization} \\
\midrule
Mistral-7B-v0.3        &57.18 &53.99 &50.27 \\
0.10 & 56.80 &59.86 & 50.92 \\
0.20 & 59.43 &62.01 &51.87 \\
0.30 & 61.99 & 64.11 & 52.17 \\
0.35 & 63.01 & 64.96 & \bf52.41 \\
0.40 & 63.88 & 66.03 & 52.11 \\
0.45 & \bf64.07 & 66.55 & 51.55 \\
0.50 & 63.57 & 66.57 & 51.39 \\
0.55 & 63.24 & \bf66.86 & 50.73 \\
0.60 & 62.38 & 66.30 & 49.79 \\
0.70 & 59.65 & 64.77 & 47.72 \\
0.80 & 56.42 & 62.78 & 46.53 \\
0.90 & 49.71 & 60.36 & 46.67 \\
\bottomrule
\end{tabular}
\label{tab:sensitivity_to_lambda}
\end{table*}

\section{Inference Efficiency Analysis}
Table \ref{tab:inference_efficiency} presents the computational cost breakdown for both Transformer and MLP architectures in terms of FLOPs per token. As demonstrated, the primary difference in computational efficiency stems from the absence of attention mechanisms in pure MLP models.

\begin{table*}[h]
\centering
\caption{Flops per Token at inference time. Following \cite{scaling_law_openai}, we analyze computational requirements for Transformer and MLP architectures where $n_{layer}$(number of layers), $d_{model}$(dimension of the residual stream), $d_{ff}$(dimension of the intermediate feed-forward layer), $d_{attn}$(dimension of the attention output) , $n_{heads}$(number of attention heads per layer), $n_{ctx}$(the length of input context), $n_{vocab}$(vocabulary size). $C_{forward}$ denotes computational cost per token inference step.
}
\resizebox{\textwidth}{!}{
\begin{tabular}{lc|c|c}
\toprule
& Openration & FLOPs per Token(Transformer) & FLOPs per Token(MLP) \\  \midrule
&Embed   & $4d_{model}$ & $-$  \\
&Attention: QKV & $2n_{layer}d_{model}3d_{attn}$ & $-$ \\
&Attention: Mask  & $2n_{layer}n_{ctx}d_{attn}$ & $-$\\
&Attention: Project& $2n_{layer}d_{attn}d_{model}$ & $-$\\
&Feedforward & $2n_{layer}2d_{model}d_{ff}$ & $3n_{layer}2d_{model}d_{ff}$\\
&De-embed  & $2d_{model}n_{vocab}$ & $2d_{model}n_{vocab}$ \\ \midrule

&Total(Non-Embedding) & $C_{forward} =4n_{layer}d_{model}(2d_{attn} + d_{ff})$ & $C_{forward}=6n_{layer}d_{model}d_{ff}$ \\ & &$+ 2n_{layer}n_{ctx}d_{attn}$\\

\bottomrule
\end{tabular}}
\label{tab:inference_efficiency}
\end{table*}



By comparing these computational requirements, we derive the theoretical speed ratio between the Transformer (denoted as $FLOPs_t$) and the MLP models (denoted as $FLOPs_m$):
\begin{equation}
\begin{split}
    \frac{FLOPs_t}{FLOPs_m} \approx \frac{4n_{layer}d_{model}(2d_{attn} + d_{ff}) + 2n_{layer}n_{ctx}d_{attn}}{6n_{layer}d_{model}d_{ff}} = 1 + \frac{n_{ctx}}{12d_{model}}, \\ \quad with \ the \ standard \quad d_{attn} = d_{ff} /4 = d_{model}.
\end{split}
\label{eq:inference_speed}
\end{equation}

This relationship in Equation \ref{eq:inference_speed} reveals that MLPs maintain a consistent computational advantage across all context lengths, with the efficiency gap widening as context length increases.

\section{Comparing Output Distribution Characteristics of LM, $k$NN, and MLP Memory}
\begin{figure}[h]
  \centering
  \begin{subfigure}[b]{0.48\textwidth} 
    \centering
    \includegraphics[width=\textwidth]{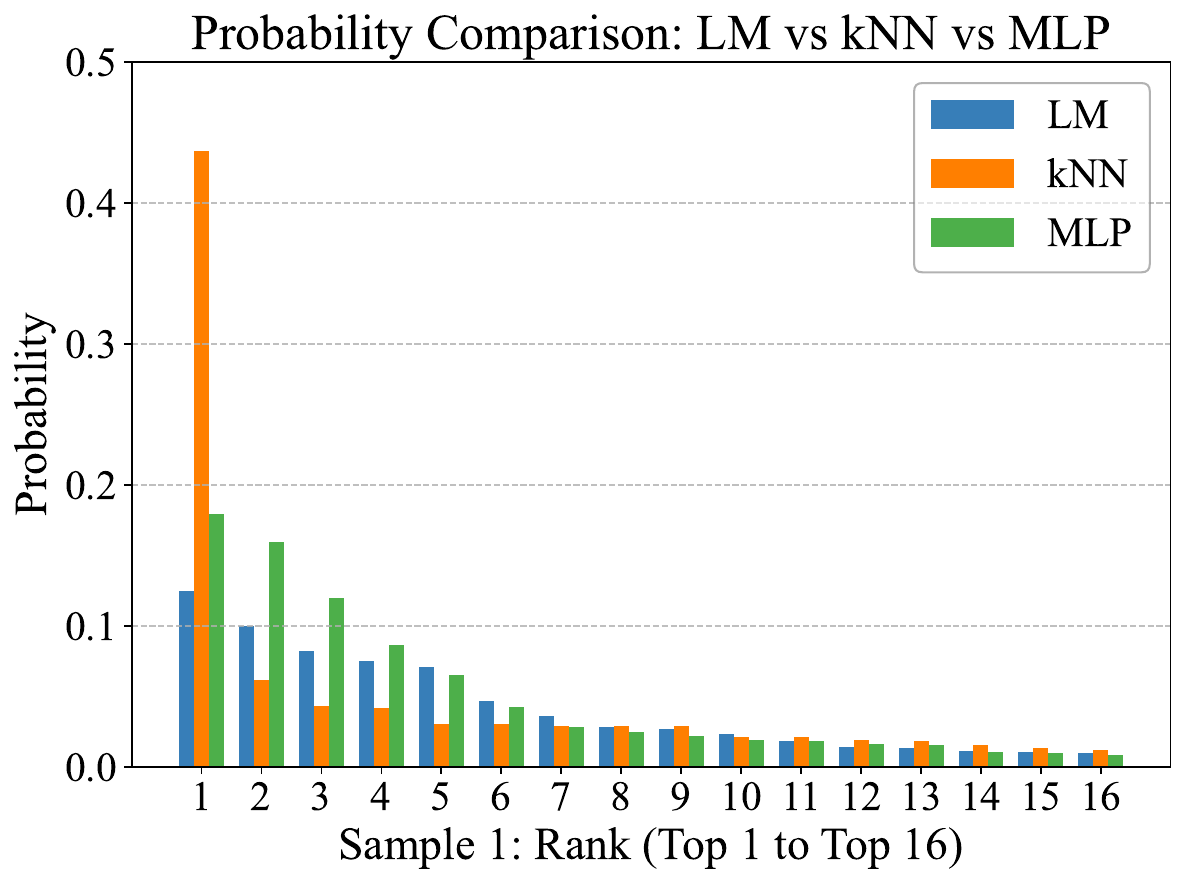}
  \end{subfigure}
  \hfill 
  \begin{subfigure}[b]{0.48\textwidth}
    \centering
    \includegraphics[width=\textwidth]{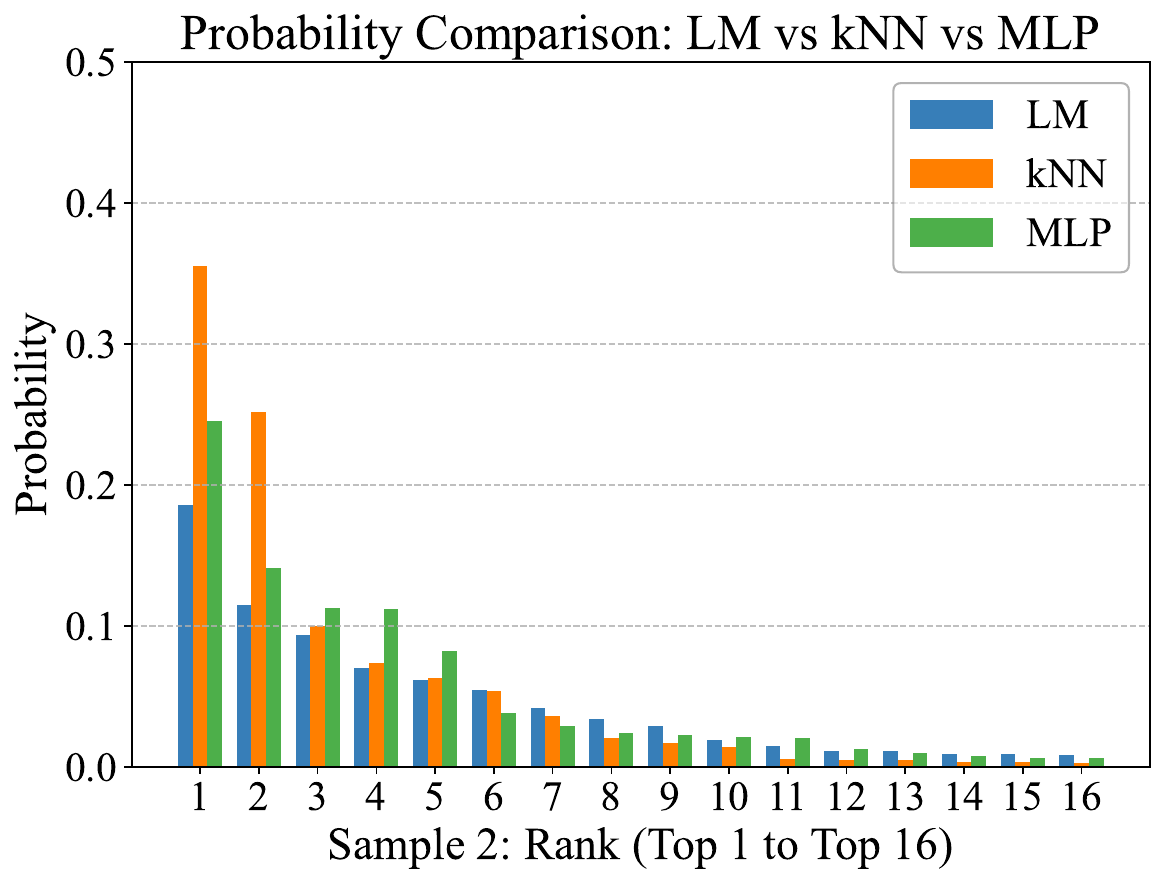}
  \end{subfigure}

  \caption{Comparison of output probability distributions. Two samples show the top-16 probabilities from the LM and $k$NN distributions using GPT2-large, along with the distribution generated by the MLP Memory based on the same large model size.
  }
  \label{fig:different_distribution}
\end{figure}

As two samples illustrated in Figure \ref{fig:different_distribution}, distributions produced by LM, $k$NN search, and MLP Memory exhibit distinct characteristics. LM typically yields smooth and dense probability distributions, as it is trained to generalize across large corpora and capture broad contextual patterns.

In contrast, the kNN approach produces sparse and spiky distributions, concentrating most of the probability mass on only a few retrieved neighbors. For instance, when using a GPT-2 model (vocabulary size 50,257), even after retrieving the top-$k$ neighbors (e.g., $k=1024$), only a small subset of these neighbors meaningfully influences the output distribution, while the majority receive near-zero probability.

The MLP Memory lies between LMs and kNN in terms of distribution characteristics. As a neural model, it is trained using a combination of KL loss and CE loss to approximate the sparse and spiky distributions produced by the kNN approach. While its outputs remain somewhat smoother due to the training objective, the resulting distributions are sharper than those of standard LMs, yet not as extreme as the highly concentrated outputs of kNN.

\begin{table*}[h]
\centering
\caption{Number of tokens with non-zero probability mass at different thresholds. This table reports the number of tokens assigned non-zero probabilities by the LM, $k$NN, and MLP Memory, across a range of probability thresholds. All values are averaged over 20,000 test samples.
}
\resizebox{\textwidth}{!}{
\begin{tabular}{lccccccc}
\toprule
 Types& $>0.0$ & $ > 10^{-6}$ & $> 10^{-5}$ & $> 10^{-4}$ & $> 10^{-3}$ & $> 10^{-2}$ & $> 10^{-1}$\\ \midrule
LM   & 50257 & 1760 &562 & 148 & 34 & 7 & 2 \\ \midrule
$k$NN & 251 & 217 & 197 & 146 & 43 & 9 & 2\\ \midrule
MLP & 50257 & 1151 & 388 & 115 & 30 & 7 & 2\\

\bottomrule
\end{tabular}}
\label{tab:statics}
\end{table*}

Table \ref{tab:statics} compares the output sparsity of LM, $k$NN, and MLP Memory by reporting the number of tokens assigned non-zero probabilities at various thresholds. The LM assigns non-zero mass to all 50,257 tokens, reflecting its dense distribution. However, this number drops sharply at higher thresholds, with only 2 tokens receiving probabilities above 0.1, indicating a rapid decay despite its broad support.

In contrast, the $k$NN output is highly sparse, with only 251 tokens assigned any non-zero probability. Even at low thresholds (e.g., $10^{-6}$), the number remains limited, confirming its concentrated nature shaped by a small set of retrieved neighbors.

MLP Memory exhibits intermediate behavior. Although it outputs over the full vocabulary like the LM, the number of tokens exceeding higher thresholds aligns more closely with $k$NN. This suggests that MLP Memory learns to approximate the spiky distributions of $k$NN while maintaining some smoothness from its parametric formulation.

\begin{table*}[h]
\centering
\caption{Cumulative token count required to reach probability mass thresholds. This table indicates the number of top-ranked tokens needed to accumulate a total probability mass exceeding thresholds such as 0.8, 0.9, etc. All values are averaged over 20,000 test samples.
}
\resizebox{\textwidth}{!}{
\begin{tabular}{lcccccc}
\toprule
 Types  & Top Prob Count(sum $>$ 0.8) & sum $>$ 0.9 & sum  $>$ 0.95 & sum  $>$ 0.99\\ \midrule
LM   & 23 & 63 & 142 & 617 \\ \midrule
$k$NN & 22 & 43 & 68 & 126\\ \midrule
MLP & 13 & 33 & 72 & 308\\

\bottomrule
\end{tabular}}
\label{tab:statics_v2}
\end{table*}

Table \ref{tab:statics_v2} further examines distribution sharpness by reporting the number of top-ranked tokens needed to accumulate a specified proportion of total probability mass. Here, we observe that the $k$NN distribution reaches 99\% cumulative probability with only 126 tokens, while the language model (LM) requires 617 tokens to achieve the same threshold. This suggests that the LM's probability mass is more broadly spread across the vocabulary, in contrast to the highly concentrated outputs of $k$NN.

Interestingly, the MLP Memory achieves 99\% cumulative probability with 308 tokens, placing it between LM and $k$NN. Notably, the MLP reaches 80\% total probability with only 13 tokens—fewer than both LM and $k$NN—indicating that it captures prominent signals more efficiently. These results support the observation that MLP Memory produces sharper distributions than LMs, yet avoids the extreme sparsity of $k$NN.

\section{Effect of Different $k$NN Target Distributions}

\begin{figure}[h]
\centering
\includegraphics[width=\textwidth]{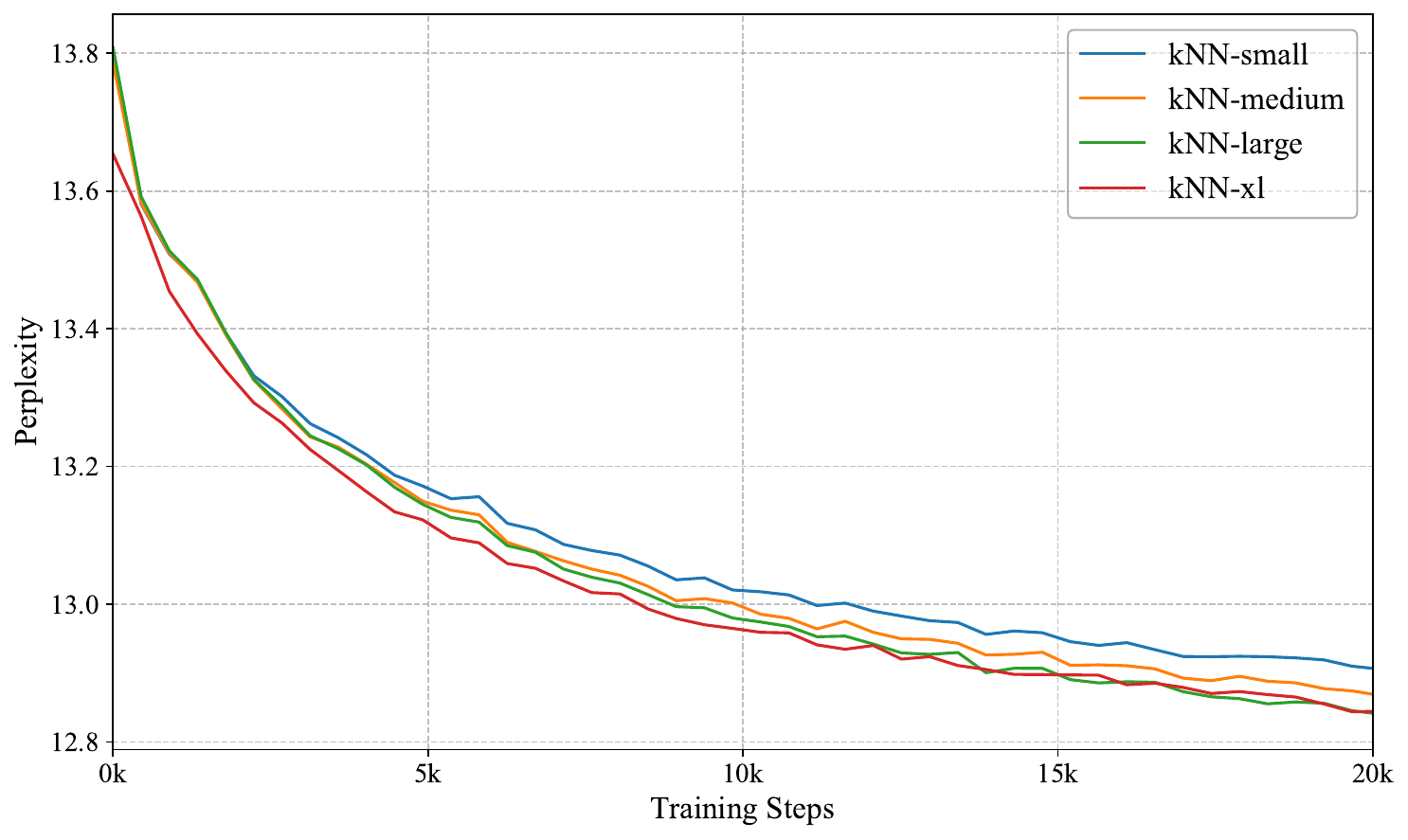}
\caption{Test perplexity of our overall model architecture, where both the base language model and the MLP Memory are of small size (GPT2-small). The MLP Memory is trained to mimic different $k$NN target distributions constructed from various base models: $k$NN-small (GPT2-small), $k$NN-med (GPT2-medium), $k$NN-large (GPT2-large), and $k$NN-XL (GPT2-xl).}
\label{fig:different_knn_target}
\end{figure}

Figure~\ref{fig:different_knn_target} presents the test perplexity of our overall model architecture evaluated at various training steps. In all settings, both the base language model and the MLP Memory are of small size (GPT2-small), with the MLP Memory trained to mimic $k$NN target distributions constructed from different base models: GPT2-small, GPT2-medium, GPT2-large, and GPT2-xl. As training progresses, test perplexity steadily declines across all variants, indicating stable optimization and effective learning. Among them, the model trained on the $k$NN-XL distribution achieves the lowest final test perplexity (12.84), closely followed by the one trained on $k$NN-large (12.85). In contrast, the models trained on $k$NN-medium and $k$NN-small converge to higher perplexities of approximately 12.87 and 12.91, respectively.

These results demonstrate that $k$NN target distributions derived from larger base models lead to improved performance when used to train the MLP Memory. The richer and more informative supervision encoded in these distributions appears to enhance the parametric memory’s generalization ability.

\section{Sensitivity to $k$ in Target Distribution Generation}
We used $k$ = 1024 for generating all target distributions. Table~\ref{tab:k_sensitivity} shows the sensitivity analysis using GPT2-large-CPT, namely GPT2-large with continued pre-training on WikiText-103. While smaller $k$ values degrade performance, values beyond 1024 yield minimal gains while significantly increasing computational costs, making $k$ = 1024 optimal for practical deployment.

\begin{table}[h]
\centering
\caption{Test perplexity sensitivity to different values of $k$ in target distribution generation using GPT2-large-CPT on WikiText-103.}

\begin{tabular}{lcc}
\toprule
Models & $k$ & Perplexity\\ \midrule
GPT2-large-CPT   &  $-$    & 10.43  \\
\midrule
\multirow{11}{*}{+$k$NN-LM}
& 1 & 10.30 \\
& 2 & 10.11 \\
& 4 & 9.95 \\
& 8 & 9.83 \\
& 16 & 9.71 \\
& 32 & 9.63 \\
& 64 & 9.57 \\
& 128 & 9.52 \\
& 256 & 9.48 \\
& 512 & 9.46 \\
& 1024 & 9.43 \\
& 2048 & 9.42 \\

\bottomrule
\end{tabular}
\label{tab:k_sensitivity}
\end{table}

\section{Case Study on Downstream Tasks}

\begin{figure}[t]
\centering
\includegraphics[width=\textwidth]{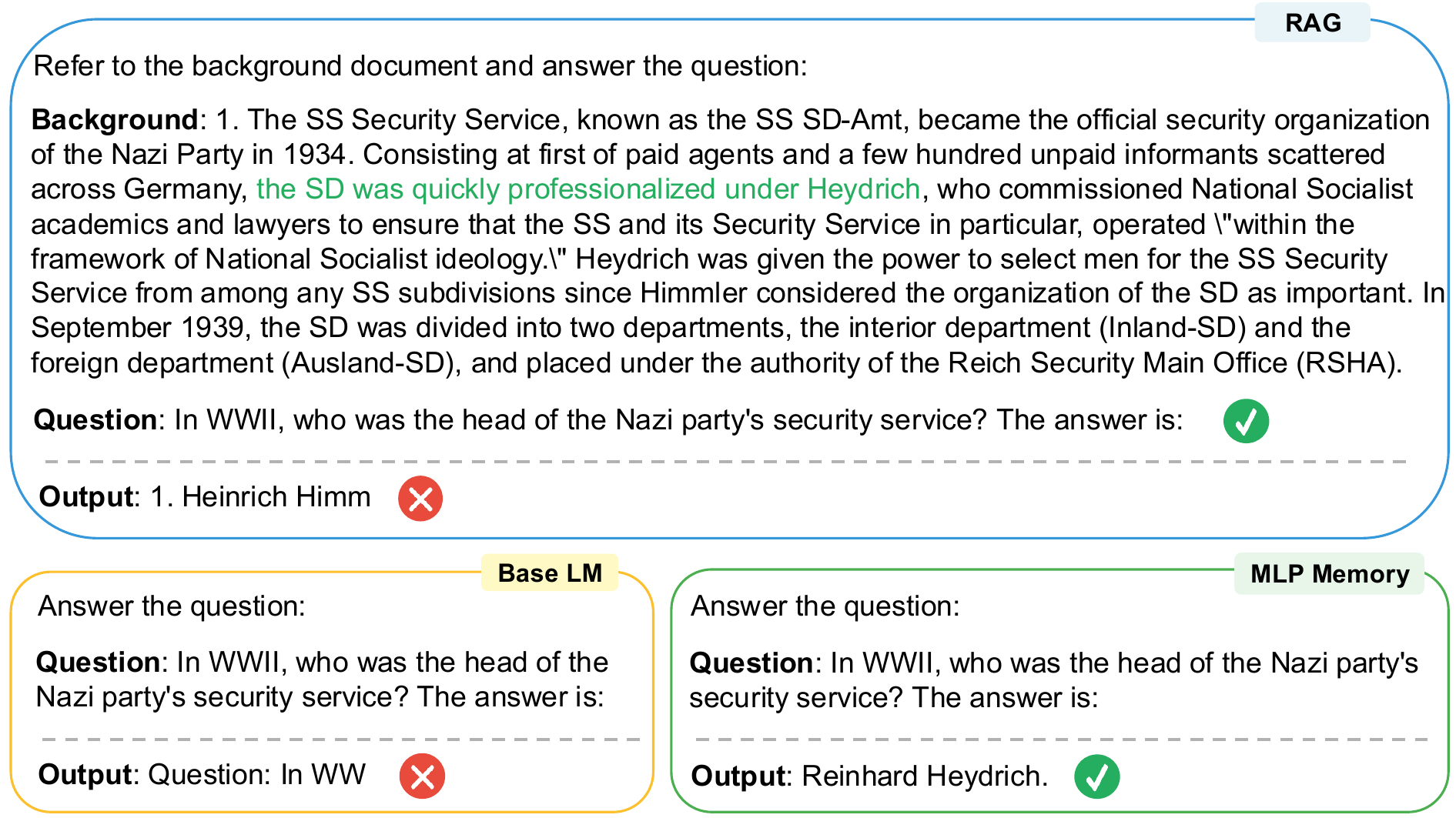}
\caption{Comparison of model outputs on a factual question. Despite retrieving relevant documents with correct information (highlighted in green), RAG is misled by contextual distractors and produces an incorrect answer. MLP Memory generates the correct answer without explicit retrieval.}
\label{fig:case_study}
\end{figure}

As show in Figure~\ref{fig:case_study}, we observe that RAG often fails even with relevant retrieved documents due to contextual noise interference. When documents contain related but distracting information, RAG's shallow integration cannot effectively filter these distractors, leading to incorrect answers. In contrast, MLP Memory learns intelligent corpus compression during training, capturing richer contextual relationships that enable robust disambiguation without explicit retrieval.

\newpage

\begin{figure}[t]
\centering
\includegraphics[width=\textwidth]{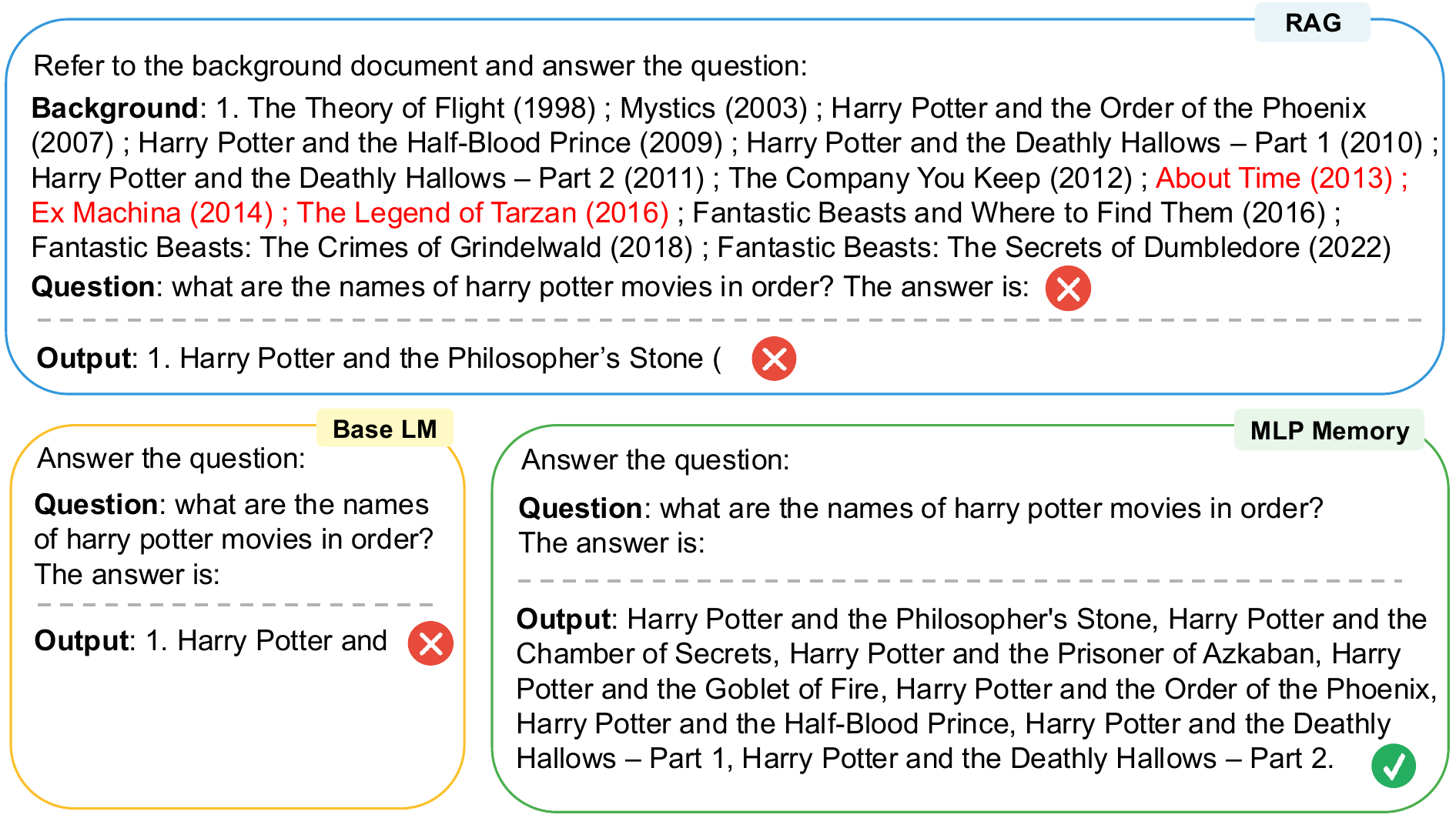}
\caption{RAG retrieves irrelevant documents that introduce interference, while MLP Memory demonstrates perfect performance.}
\label{fig:example1}
\end{figure}

\begin{figure}[t]
\centering
\includegraphics[width=\textwidth]{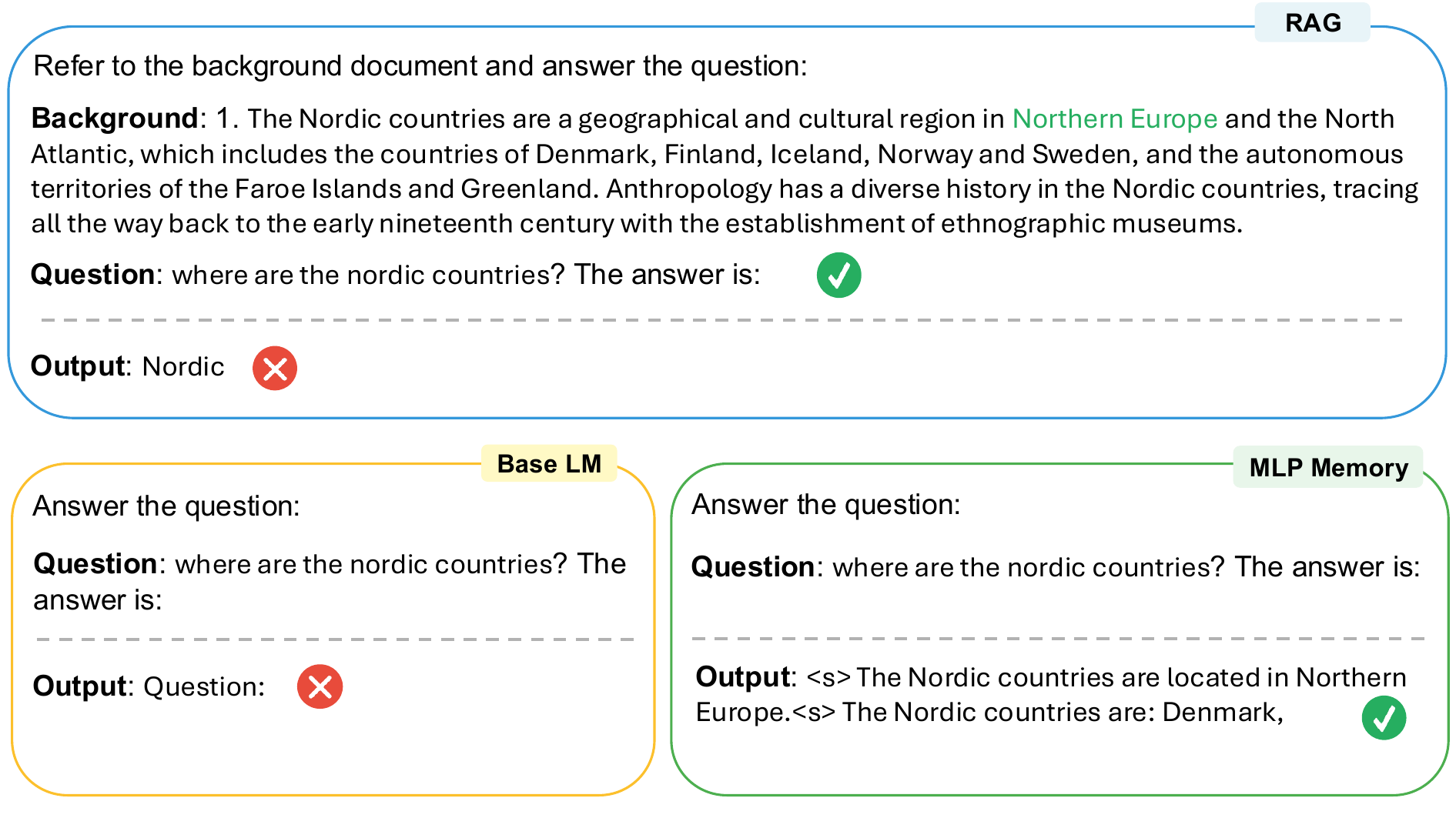}
\caption{RAG fails to extract answer despite retrieving relevant content, while MLP Memory provides accurate response}
\label{example2}
\end{figure}

\begin{figure}[t]
\centering
\includegraphics[width=\textwidth]{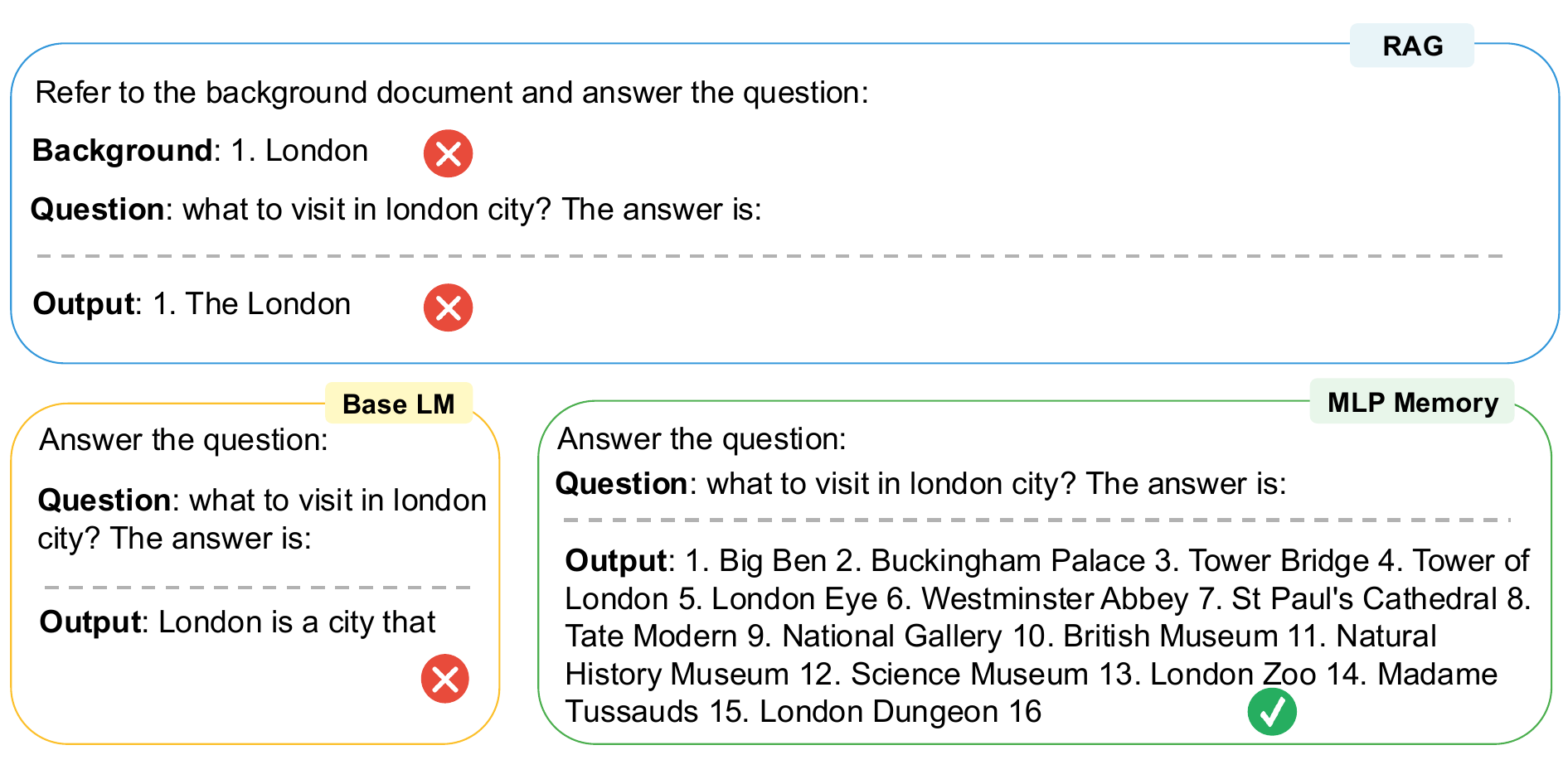}
\caption{RAG system distracted by retrieved content fails to answer the question, while MLP Memory provides comprehensive and accurate response}
\label{fig:example3}
\end{figure}

\clearpage

\section{Case Study on the Distribution of Base LM, $k$NN-LM, and MLP Memory}
\label{app:case_study_distribution}

To further understand the mechanisms underlying the effectiveness of MLP Memory, we analyze the token-level probability distributions produced by the base LM, kNN-LM, and MLP Memory. Our hypothesis is that kNN-based distributions are particularly effective at capturing long-tail knowledge, whereas LM distributions exhibit strong coherence on frequent function words. MLP Memory appears to internalize both characteristics, acquiring long-tail information in a manner similar to non-parametric retrieval while preserving the stability of parametric models.

To examine this hypothesis, we perform case studies on examples drawn from WikiText-103 and report the probability assigned by each method to the target token (highlighted in bold) given its preceding context.

\begin{table}[h]
\centering
\small
\caption{Token-level probability assignments for different methods on long-tail entities (\colorbox{SkyBlue!20}{\textit{top block}}) and coherent function words (\colorbox{Apricot!20}{\textit{bottom block}}).}
\resizebox{\linewidth}{!}{
\begin{tabular}{p{8.5cm}ccc}
\toprule
\rowcolor{SkyBlue!20}
\multicolumn{4}{c}{\textbf{Long-tail Knowledge}} \\
\midrule
\textbf{Context (target token in bold)} & \textbf{Base LM} & \textbf{kNN-LM} & \textbf{MLP Memory} \\
Southward, in the Yongsan area, Keiser placed Brigadier General Joseph S.\ \textbf{Bradley}, Assistant Division Commander, in charge of the 9th Infantry Regiment. &
0.01 & 0.74 & \textbf{0.75} \\[4pt]

The song reached number ten in \textbf{Mexico} and number one on both the Billboard Latin Songs and Latin Pop Songs chart. &
0.01 & 0.07 & \textbf{0.45} \\
\midrule
\rowcolor{Apricot!20}
\multicolumn{4}{c}{\textbf{Coherence}} \\
\midrule
\textbf{Context (target token in bold)} & \textbf{Base LM} & \textbf{kNN-LM} & \textbf{MLP Memory} \\
As the threat of invasion was clearly felt in late 1941, an idea for a series of secret observation posts (first in Gibraltar and later \textbf{in} other places like Malta and Aden)... &
\textbf{0.65} & 0.01 & 0.44 \\[4pt]

Here the invasion force encountered the first French defences, consisting of camouflaged trenches \textbf{and} pillboxes dug in along a ridge. &
0.45 & 0.06 & \textbf{0.53} \\
\bottomrule
\end{tabular}
}
\label{tab:distribution_case_study}
\end{table}

As shown in Table~\ref{tab:distribution_case_study}, the case studies provide clear empirical support for our hypothesis. For rare entities such as \textit{Bradley} and \textit{Mexico}, MLP Memory assigns probabilities comparable to or exceeding those of kNN-LM, demonstrating effective acquisition of long-tail knowledge. In contrast, for function words such as \textit{in} and \textit{and}, MLP Memory maintains probability mass close to that of the base LM, whereas kNN-LM shows substantial degradation. These observations suggest that MLP Memory successfully combines the advantages of non-parametric retrieval with the coherence properties of parametric language models.

\section{MLP Architecture Details}
\label{mlp_architecture}

The MLP Memory used in LLaMA2- and Mistral-based experiments is initialized from the corresponding MLP modules of their original architectures to maintain structural consistency and stable training behavior. Both settings employ 8 stacked MLP layers with their respective native hidden and intermediate dimensions. The total size of the MLP Memory is about 1B parameters, excluding embedding parameters. The detailed architectural configurations are reported in Table~\ref{table:mlp_arch_backbones}.

\begin{table}[ht!]
    \centering
    \caption{MLP Memory architecture for LLaMA2 and Mistral experiments.}
    \begin{tabular}{ccc}
        \toprule
     \textbf{Layers} & \textbf{Hidden dim} & \textbf{Intermediate dim} \\
        \midrule
         8 & 4096 & 11008 \\
         8 & 4096 & 14336 \\
        \bottomrule
    \end{tabular}
    \label{table:mlp_arch_backbones}
\end{table}

In the study of the effect of MLP Memory size presented in Appendix \ref{mlp_memory_size}, the number of layers is varied while the hidden and intermediate dimensions are fixed at 1280 and 5120, respectively. The corresponding configurations are summarized in Table~\ref{table:mlp_ablation}.

\begin{table}[ht!]
    \centering
    \caption{MLP Memory size ablation configurations.}
    \begin{tabular}{ccc}
        \toprule
        \textbf{Layers} & \textbf{Hidden dim} & \textbf{Intermediate dim} \\
        \midrule
        8  & 1280 & 5120 \\
        15 & 1280 & 5120 \\
        36 & 1280 & 5120 \\
        \bottomrule
    \end{tabular}
    \label{table:mlp_ablation}
\end{table}

Based on the MLP Memory size study, we adopt the 8-layer configuration as the default setting, as it provides a favorable balance between performance and efficiency.

\section{Effect of Different MLP Memory Sizes on Performance}
\label{mlp_memory_size}

This section examines how different MLP Memory sizes affect language modeling performance on WikiText-103. The memory capacity is controlled by varying the number of MLP layers while keeping other architectural settings unchanged.

\begin{table}[ht!]
    \centering
    \caption{Performance of different MLP Memory sizes on WikiText-103. The base model is GPT2-large.}
    \begin{tabular}{lc}
        \toprule
        \textbf{Model} & \textbf{WikiText-103 PPL} \\
        \midrule
        GPT2-large & 15.81 \\
        + MLP Memory (8 layers, 221M)  & 11.41 \\
        + MLP Memory (15 layers, 359M) & 11.35 \\
        + MLP Memory (36 layers, 772M) & 11.25 \\
        \bottomrule
    \end{tabular}
    \label{table:mlp_size}
\end{table}

As shown in Table~\ref{table:mlp_size}, even the smallest MLP Memory achieves significant improvements over the base model. While scaling provides additional gains, smaller models offer the best performance-efficiency trade-off.

\end{document}